\documentclass[conference]{IEEEtran}
\usepackage{times}
\usepackage{amssymb}
\usepackage{mathtools}
\usepackage{algorithm}
\usepackage{algpseudocode}
\usepackage{xcolor}
\usepackage{booktabs}
\usepackage{enumitem}
\usepackage{bbm}
\usepackage{subfig}

\usepackage[numbers]{natbib}
\usepackage{multicol}
\usepackage[bookmarks=true]{hyperref}

\usepackage{xcolor}
\usepackage{xspace}
\usepackage{bbm}
\usepackage{makecell}

\definecolor{orange(sae/ece)}{rgb}{1.0, 0.49, 0.0}
\definecolor{teal(sae/ece)}{rgb}{0, 0.47, 0.52}
\definecolor{purple}{rgb}{0.74, 0.65, 1.0}
\definecolor{dark_purple}{rgb}{0.72, 0.33, 0.82}
\definecolor{light_gray}{rgb}{0.9, 0.9, 0.9}
\definecolor{medium_gray}{rgb}{0.6, 0.6, 0.6} 
\definecolor{dark_gray}{rgb}{0.2, 0.2, 0.2} 
\definecolor{dark_blue}{rgb}{0.098, 0.239, 0.52}
\definecolor{dark_brown}{rgb}{0.3255, 0.004, 0.001}
\definecolor{r3mcolor}{rgb}{0.478, 0.1569, 0.4863}
\definecolor{light_blue}{rgb}{0.33, 0.80, 1}

\newcommand{\acqr}{\textcolor{orange}{\textbf{ACQR}}\xspace}
\newcommand{\qr}{\textcolor{black}{\textbf{QR}}\xspace}
\newcommand{\ensemble}{\textcolor{gray}{\textbf{Ensemble}}\xspace}

\newcommand{\edit}[1]{{\color{black}{#1}}}

\newcommand{\para}[1]{\smallskip \noindent \textbf{{#1}.}}

\newcommand{\human}{\HH}

\newcommand{\state}{s} 
\newcommand{\stateSpace}{\mathcal{S}}
\newcommand{\dyns}{f}

\newcommand{\aH}{a_{\human}} 
\newcommand{\aHhat}{\hat{a}_{\human}} 
\newcommand{\aR}{a} 
\newcommand{\actionSpace}{\AA}
\newcommand{\lowH}{u_{\human}} 
\newcommand{\inputSpace}{\HH}

\newcommand{\controller}{f}

\newcommand{\dtrain}{\DD_\mathrm{train}}
\newcommand{\dcalib}{\DD_\mathrm{calib}}
\newcommand{\estqlo}{\hat{q}_{\alpha_{lo}}}
\newcommand{\estqhi}{\hat{q}_{\alpha_{hi}}}

\newcommand{\errt}{\text{err}_t}

\def\RR{\mathbb{R}}

\def\AA{\mathcal{A}}

\def\HH{\mathcal{H}}

\def\DD{\mathcal{D}}

\newlist{SL}{enumerate}{1}
\setlist[SL]{label=\textbf{S\arabic*:}}

\begin{document}

\title{Conformalized Teleoperation: Confidently Mapping Human Inputs to High-Dimensional Robot Actions}

\author{Michelle Zhao,
Reid Simmons,
Henny Admoni, 
Andrea Bajcsy\\ Robotics Institute, Carnegie Mellon University\\
\{mzhao2, rsimmons, hadmoni, abajcsy\}@andrew.cmu.edu}

\maketitle

\begin{abstract}
Assistive robotic arms often have more degrees-of-freedom than a human teleoperator can control with a low-dimensional input, like a joystick. To overcome this challenge, existing approaches use data-driven methods to learn a mapping from low-dimensional human inputs to high-dimensional robot actions. However, determining if such a black-box mapping can \textit{confidently} infer a user's \textit{intended} high-dimensional action from low-dimensional inputs remains an open problem. Our key idea is to adapt the assistive map at training time to additionally estimate high-dimensional action quantiles, and then calibrate these quantiles via rigorous uncertainty quantification methods. Specifically, we leverage adaptive conformal prediction which adjusts the intervals over time, reducing the uncertainty bounds when the mapping is performant and increasing the bounds when the mapping consistently mis-predicts. Furthermore, we propose an uncertainty-interval-based mechanism for detecting high-uncertainty user inputs and robot states. We evaluate the efficacy of our proposed approach in a 2D assistive navigation task and two 7DOF Kinova Jaco tasks involving assistive cup grasping and goal reaching. Our findings demonstrate that conformalized assistive teleoperation manages to detect (but not differentiate between) high uncertainty induced by diverse preferences and induced by low-precision \edit{trajectories} in the mapping's training dataset. On the whole, we see this work as a key step towards enabling robots to quantify their own uncertainty and proactively seek intervention when needed. 

\end{abstract}

\IEEEpeerreviewmaketitle

\section{Introduction}
\label{sec:introduction}

Robotic arms designed for assistance typically possess more degrees of freedom (DoF) than can easily be controlled by a human using low-dimensional inputs such as a joystick \cite{tsui2008development} or a sip-and-puff device \cite{boboc2012review, javdani2018shared}. 
In practice, teleoperating a high-DoF robot can require frequent mode switches that control individual components of the arm during even very basic everyday tasks \cite{eftring1999technical, rosier1991rehabilitation, tijsma2005framework}. 
For this reason, a core challenge in assistive teleoperation is building mappings that translate low-dimensional human inputs into the human's desired high-dimensional robot actions in an intuitive way \cite{losey2020controlling}. 

Recently, these low-to-high dimensional mappings have been increasingly learned from human data, wherein annotators provide state, high-DoF action, and sometimes low-DoF input pairs for training \cite{losey2020controlling, li2020learning, karamcheti2022lila}. 

However, quantifying the trustworthiness of these black-box assistive teleoperation systems is an open problem. We find that off-the-shelf, these learned mappings can be highly sensitive to multimodality and suboptimal data in the training distribution. 
Furthermore, when faced with out-of-distribution human inputs at deployment time, the robot can overconfidently map the human's low-DoF \edit{input} to an extremely incorrect high-DoF robot action. 
At best, this can cause frustration from the end-user; at worst, it can be a safety hazard. 

\begin{figure}[t]
    \centering
    \includegraphics[width = 0.49\textwidth]{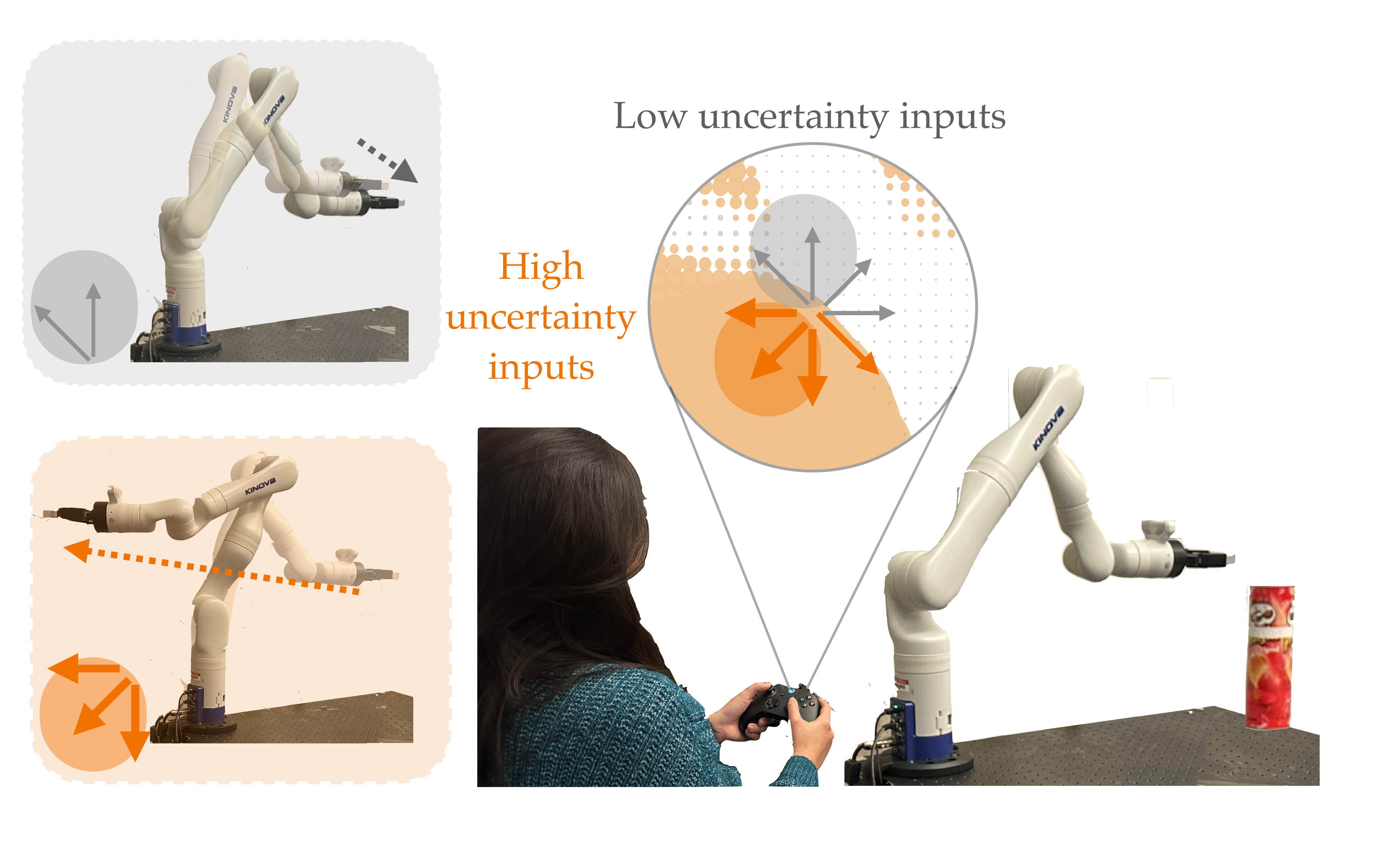}
    \caption{\textbf{Conformalized Teleoperation.} We leverage conformal methods to quantify if the robot's learned controller can reliably lift the human's low-DoF input (joystick) to their desired high-DoF action (7 joint velocities).
    For any joystick input at the current state, the robot can assess its uncertainty in remapping that input (dot size in the expanded view is proportional to uncertainty at that coordinate). 
    Arrows emphasize directional joystick input.
    (left, top) If the human pushes \textit{up} or to the \textit{\edit{left}} on the joystick, the robot has low uncertainty, because it knows with high probability the person wants to go forward \edit{towards the object}. 
    (left, bottom) If the human pushes \textit{backwards} on the joystick, the robot predicts a large pivot backwards, but is rightfully uncertain this is what the human intended.
     }
    \label{person_using}
\end{figure}

In this work, we seek to rigorously quantify 
if a learned assistive controller can \textit{confidently} infer a user's \textit{intended} high-dimensional action from only low-dimensional states and inputs. 
Our key idea is: 
\begin{quote}
    \textit{by training the robot's controller to estimate high-DoF action quantiles, we can calibrate the uncertainty via adaptive conformal prediction techniques.}
\end{quote}

Specifically, we leverage the statistical technique of adaptive conformalized quantile regression \cite{gibbs2021adaptive}, which is relevant for regression problems and provides asymptotic coverage \edit{guarantees} (i.e., human's true high-DoF action is included within the uncertainty interval) without strong distributional assumptions. 
Furthermore, the adaptive nature of this paradigm enables the calibration procedure to  grow and shrink uncertainty intervals over time as the end-user's input is poorly or well-predicted by the robot's learned controller.
As such, we call our approach Conformalized Teleoperation. 
With this rigorous quantification of the robot's uncertainty, we also propose a simple method to detect high-uncertainty states which can be used to alert the end-user or safely stop operation.

We evaluate how well our approach calibrates in a 2D assistive navigation task, and two 7DOF Kinova Jaco tasks involving assistive cup grasping and goal reaching. 
Our experiments indicate that naturally-occurring diversity in the training dataset, such as multiple preferences for how a task should be completed or low-precision behaviors, can significantly impact the robot's learned assistive control confidence.  
Beyond this, we find that our approach is capable of detecting critical high-uncertainty states and inputs for both in- and out-of-distribution end user behavior.
Overall, we see this work as an important first step towards robots that rigorously quantify their own uncertainty in the assistive teleoperation domain. Project website is available at \url{https://conformalized-assistive-teleoperation.github.io/}.

\section{Related Work}
\label{sec:related_work}

\para{Assistive Teleoperation} Assistive robots enable users with physical impairments to perform a variety of everyday tasks \cite{brose2010role, miller2006assistive}. Teleoperating a high-DoF robot can be difficult and often requires frequent mode switches to accomplish simple tasks \cite{eftring1999technical, rosier1991rehabilitation, tijsma2005framework}. One of the key challenges is the low-to-high-dimensional mapping induced by low-DoF controllers like joysticks \cite{tsui2008development}, sip-and-puff devices \cite{boboc2012review, javdani2018shared}, or eye gaze \cite{aronson2020eye}. 
Prior works on dimensionality reduction have shown increasingly sophisticated methods for designing or learning a mapping from the user's input to the robot's high-DOF control action \cite{losey2020controlling, jeon2020shared, losey2022learning, gopinath2020active}. 
Alternatively, the shared control paradigm combines low-DoF human input with autonomous robot policies in order to make teleoperation of the robot more fluent \cite{broad2020data, broad2018learning}. 
Other works considers the inverse approach of developing high-dimensional interfaces to control high-DoF robots \cite{lee2021learning}. In this work, we draw upon the data-driven approaches \cite{losey2020controlling, losey2022learning} but focus on getting statistical performance assurances on the output of the model \cite{angelopoulos2021gentle, gibbs2021adaptive} when the assistive mapping is a neural network.

\para{Uncertainty-Aware Shared Control}
In shared control paradigms, robots must distinguish intents of the human operator \cite{broad2020data}. 
Uncertainty-aware shared control paradigms can elect to prioritize human control, handing off control to the human operator, while constraining robot inputs to maintain safety \cite{broad2018learning}, or suggesting a control mode in which human input maximally disambiguates their intent \cite{gopinath2020active}. Prior works maintain an estimate of intent confidence for low-dimensional hand-designed intent spaces \cite{zurek2021situational, dragan2013policy} or, for data-driven estimators, when relying on data-driven based methods like learned encoder-decoder architectures, 
use reconstruction error as a signal of intent inference quality \cite{jonnavittula2022learning}.
When intent inference is detected to be poor, control is handed over to the human. In the latter category, methods assume known human intent but focus on preventing collisions during robot assistance planning \cite{you2012assisted, zhou2022cpi}. 
In a similar vein as our work, adaptive shared control approaches predict human actions online and adjust robot guidance as a result of uncertainty in the robot's inferences \cite{hara2023uncertainty}. However, we translate ideas from conformal prediction to get rigorous uncertainty bounds for learned robot assistive controllers.

\medskip 
\noindent \textbf{Uncertainty Quantification} 
seeks to estimate the confidence a model has in its predictions. 
Bayesian approaches such as Bayes filters \cite{chen2003bayesian, ching2006bayesian} and Bayesian neural networks \cite{goan2020bayesian, lampinen2001bayesian, kononenko1989bayesian} quantify uncertainty via posterior updates. 
In deep learning other popular uncertainty quantification methods include Monte Carlo dropout \cite{gal2016dropout, zhu2017deep, shamsi2021improving} and deep ensemble approaches \cite{mendes2012ensemble, lakshminarayanan2017simple, jiang2021can}. Other ensemble model disagreements can also be used such as random forests \cite{breiman2001random} and bagging \cite{breiman1996bagging}. 
We ground our approach in conformal prediction \cite{angelopoulos2021gentle}, specifically adaptive conformal inference \cite{gibbs2021adaptive}, an increasingly popular method which provides coverage guarantees without assumptions on the data distribution nor model assumptions.

\para{Conformal Prediction for Robotics} Conformal prediction is a paradigm for constructing rigorous prediction intervals for both classification and regression problems \cite{angelopoulos2021gentle, romano2019conformalized, romano2020classification, zaffran2022adaptive}. The \edit{method has been} explored in a variety of robotics contexts for providing collision-avoidance assurances \cite{chen2021reactive, lindemann2023safe, dixit2023adaptive, muthali2023multi, taufiq2022conformal, dietterich2022conformal, lin2023verification}, calibrating early warning systems \cite{luo2022sample} and large language models \cite{ren2023robots}. 
We take inspiration from these recent successes of conformal prediction techniques applied to robotics, and propose how to incorporate these advances into the assistive teleoperation domain.

\begin{figure*}[t]
    \centering
    \includegraphics[width=0.99\textwidth]{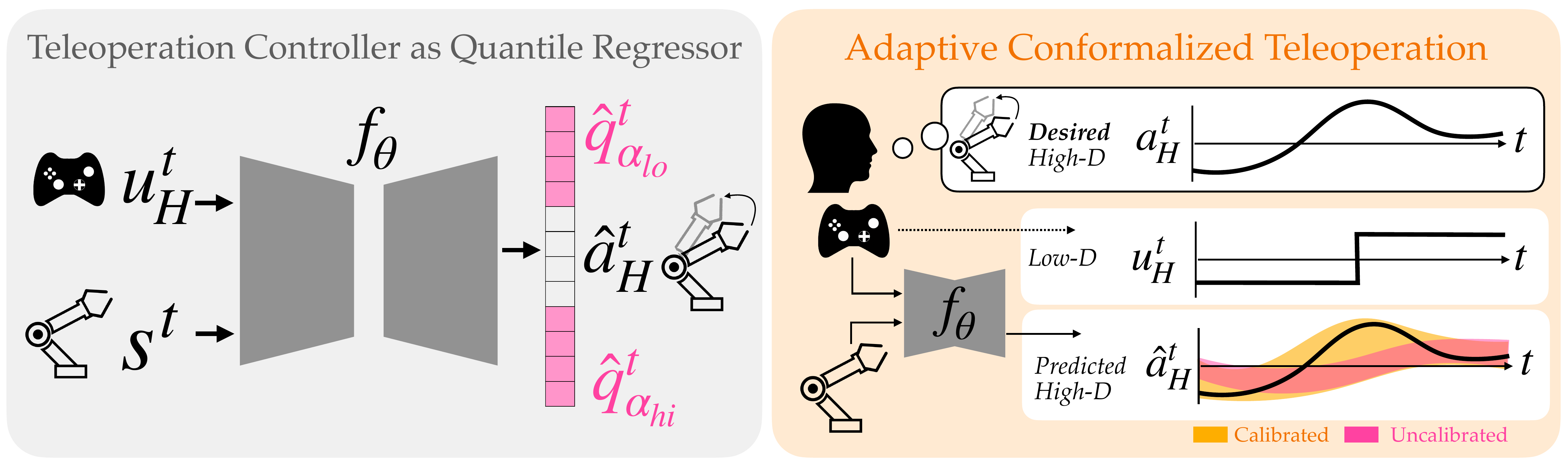}
    \caption{\textbf{Approach for Conformalized Teleoperation} (left) During training, we modify the teleoperation controller $\controller_\theta$ to regress both the high-dimensional action corresponding with the low-dimensional human input, but also the empirical quantiles. (right) When deployed around a new user, we can calibrate the model to the user's new dataset distribution. Adaptive Conformal Quantile Regression enables us to enlarge or shrink the predicted quantiles to get coverage of the user's \textit{desired} high-dimensional action.}
    \label{fig:qr_regressor}
\end{figure*}

\section{Problem Formulation}
\label{sec:problem_formulation}

\edit{In this work, we seek to rigorously quantify how \textit{confidently} a learned assistive controller can infer a user's \textit{intended} high-dimensional action from low-dimensional inputs and states. Here, we formalize the key components of our problem setup.} 

\para{Robot States \& Actions} We model the state as $\state \in \stateSpace \subseteq \RR^{n_s}$, consisting of the robot's joint angles and relevant object poses in the scene, and the robot's high-dimensional action as $\aR \in \actionSpace \subseteq \RR^{n_{a}}$ (e.g., joint velocities). The robot's state evolves via the discrete-time dynamics $\state^{t+1} = \dyns(\state^t, \aR^t)$. 

\para{Human Teleoperation} 
We model the human teleoperator as desiring to execute a specific high-DoF action on the robot: let $\aH \in \actionSpace$ be the human's desired high-dimensional action. 
However, the human teleoperates the robot via a \textit{low-dimensional} interface, $\lowH \in \inputSpace \subseteq \RR^{n_{u}}$. For example, if $n_{u} = 2$ then this could be a 2-dimensional joystick that the human uses to control a robot's 7 joint velocities, $n_{a} = 7$. 

\para{Learned Low-to-High-Dimensional Controller}
To enable easier teleoperation, we study learned assistive controllers \cite{li2020learning, losey2020controlling, karamcheti2021learning, karamcheti2022lila} which predict a human's desired high-dimensional action from the current state and low-dimensional input. Let this controller be: 
\begin{equation}
    \aHhat^t = \controller_\theta(\lowH^t, \state^t)
\end{equation}

In our setting of interest, this controller $\controller_\theta$ is represented as a neural network with parameters $\theta$ and is learned from a dataset  $(\state^t, \lowH^t, \state^{t+1}) \in \dtrain$ 
where $\state^t$ is the current state and $\state^{t+1}$ is the intended next state, and $\lowH^t$ is the corresponding low-dimensional input the human would give to yield the desired next state. 

In practice, to generate a sufficiently large and diverse training dataset, a variety of strategies can be employed: for example, using heuristics to deterministically generate $\lowH^t$ for consecutive state pairs \cite{li2020learning}, showing human annotators current and next-state robot pairs (e.g., robot is near an object, then the robot is closer to the object) and asking them to label them with the corresponding low-dimensional action (e.g., $\lowH = \mathrm{left}$ on the joystick) \cite{li2020learning}, or querying large-language models as a proxy for human labels \cite{zhang2023large}.

After training $\controller_\theta$, the robot is deployed to interact with a target user, and executes the predicted high-dimensional action $\aHhat^t$ output by the learned controller conditioned on the current state $\state^t$ and the human's low-dimensional action $\lowH^t$. 

\para{Goal: Calibrated Low-to-High Dimensional Control} 
Our goal is to calibrate the controller $\controller_\theta$ trained on $\dtrain$ to a target end-user. 
Intuitively, we want a principled measure of the learned controller's \textit{uncertainty} at different states and low-dimensional end-user inputs.
More formally, we seek to ensure that the human's \textit{intended} high-DoF action, $\aH$, is contained within the predicted action bound on $\controller_\theta$'s output with designer-specified probability $1-\alpha$.

\section{Method: Conformalized Teleoperation}
\label{sec:method}

\edit{
Our approach to quantifying the confidence of a black-box, learned assistive controller on target user data is grounded in adaptive conformal prediction \cite{gibbs2021adaptive}. While there are many variants of conformal prediction (CP) (see \cite{angelopoulos2021gentle} for an overview), we focus on Adaptive Conformalized Quantile Regression (ACQR) \cite{romano2019conformalized} which frames the problem of constructing high-confidence prediction intervals on temporally-correlated data as an online learning problem, 
is relevant to regression problems, and yields asymptotic coverage guarantees without strong distributional assumptions \cite{gibbs2021adaptive, romano2019conformalized}. 
Additional background on relevant conformal prediction concepts can be found in the Appendix, Section \ref{sec:cqr_background}. }

\edit{
In this section, we break down our Conformalized Teleoperation approach into three key steps. 
First, we train the robot's initial assistive mapping with paired low-DoF inputs and high-dimensional-DoF outputs, but force the model to approximate its own uncertainty by predicting empirical quantiles (Section~\ref{subsec:training}). 
Next, we use ACQR to calibrate this base model to data from a target user. Given some miscoverage rate $\alpha \in [0,1]$, our goal in conformal prediction to construct prediction intervals such that the probability the interval contains the correct label (i.e., action) is almost exactly $1-\alpha$. 
A target user's calibration data---drawn from the same distribution that the model will be deployed with at test time (from the target user) but unseen during training---is what enables the calibration procedure to produce prediction intervals guaranteed to contain the user's desired high-DoF action with a specified high probability (Section~\ref{subsec:acqr}). 
Finally, we use the size of the calibrated prediction intervals as a calibrated monitor indicating the model's uncertainty on a target user (Section~\ref{subsec:monitoring}). }

\subsection{Training Teleoperation Controller as a Quantile Regressor}
\label{subsec:training}
\edit{We start by training the robot's assistive teleoperation controller via a neural network, and force the model to predict its own uncertainty by predicting the empirical quantiles of the predicted high-DoF action.}

\para{Training Data} \edit{Let $\dtrain$ be a dataset} consisting of tuples $(\state^t, \lowH^t, \state^{t+1})$, where $\state^t$ is the current state, $\state^{t+1}$ is the intended next state, and $\lowH^{t}$ a low-dimensional input expected to move the robot from $\state^{t}$ to $\state^{t+1}$. 
Note that in our primary problem setting, the states are the 7DOF joint angles of the robot ($\state := q$) and actions are joint velocities ($\aR := \dot{q}$), so we can simply subtract $\state^{t}$ from $\state^{t+1}$ to determine the action $\aH^t$ intended by $\lowH^{t}$. 

\para{Architecture} 
The assistive controller $\controller_{\theta}$ is a neural network parameterized by $\theta$ that takes as as input $(\state^{t}, \lowH^{t})$ pairs and estimates $\aH^t$. 
The network architecture is a simple feed-forward neural network with GELU activation \cite{hendrycks2016gaussian}, and an information bottleneck layer of dimension 6. The information bottleneck layer is followed by a tanh activation (see Figure \ref{fig:qr_regressor}).

To construct a controller suitable for conformal inference, we train $\controller_\theta$ to additionally estimate quantiles $\estqlo=\frac{\alpha}{2}$ and $\estqhi=1-\frac{\alpha}{2}$ of the predicted action $\aHhat$ given $(\state,\lowH)$, using quantile regression \cite{koenker1978regression}. 
More formally, the $\alpha$-th conditional quantile function is
\begin{equation}
    q_{\alpha} \coloneqq \inf\{\aR \in \RR^m: P(Y \geq \aR \mid X = (\state,\lowH)) \geq \alpha \}
\end{equation}
The uncertainty in the prediction of the human's intended high-dimensional action is reflected in the size of the interval.

\para{Loss Function} We train $f_{\theta}$ to minimize:
\begin{equation}
    \mathcal{L} = \mathcal{L}_{MSE} + \mathcal{L}_{p({\alpha_{hi}})} + \mathcal{L}_{p({\alpha_{lo}})},
\end{equation}
where $\mathcal{L}_{MSE}$ measures error between the controller's predicted high-DoF action, $\aHhat$, and the desired high-DoF action, $\aH$. 

The losses, $\mathcal{L}_{p({\alpha_{hi}})}$ and $\mathcal{L}_{p({\alpha_{lo}})}$, are two pinball losses (where $p(\cdot)$ stands for pinball) that are standard within conformal \edit{literature} (as defined in \cite{koenker1978regression, romano2019conformalized, steinwart2011estimating}). \edit{Thus, in addition to the mean prediction, the model also outputs the empirical upper $1-\frac{\alpha}{2}$ and lower $\frac{\alpha}{2}$-quantile: 
\begin{equation}
    \hat{C}(\state^{t}, \lowH^{t}) = [\estqlo(\state^{t}, \lowH^{t}), \estqhi(\state^{t}, \lowH^{t})].
\end{equation}
}
For training, we use a learning rate of \edit{$0.01$}.

\subsection{Adaptive Conformalized Quantile Regression (ACQR)}
\label{subsec:acqr}

\edit{Although the base model in Section~\ref{subsec:training} is trained to estimate its own uncertainty, the estimated uncertainty interval $\hat{C}(\state^{t}, \lowH^{t})$ is not necessarily calibrated. Here, we use a calibration dataset consisting of a target user's desired low-DoF and high-DoF mappings and conformal prediction to correct $\hat{C}$ to ensure our desired coverage.}

\para{\edit{Measuring Misprediction}}\label{subsec:nonconformity_score} \edit{The critical step in ACQR \cite{gibbs2021adaptive} is to, for any input $(\state^{t}, \lowH^{t})$, compute the \textit{conformity score} that quantifies the error made by the initial prediction interval.} One \edit{design choice when} applying adaptive conformal \edit{prediction} to our robotics domain is how to scale uncertainty across the high-dimensional output vector. \edit{In general, each dimension of the high-dimensional output may not have the same scale; to account for this, we select a multiplicative (rather than additive) factor as the nonconformity score.} Specifically, we look for the smallest multiplicative factor $\rho \in \RR$ for which we get coverage across all dimensions of the output vector. 

Mathematically, for each 
datapoint $(\state^{t}, \lowH^{t}, \aH^t)$, we compute the similar upper and lower prediction error:

\edit{
\begin{equation}
\label{eq:delta_lohi}
    \Delta_t^{+} = \max\{\estqhi - \aHhat, \epsilon\}, \quad \Delta_t^{-} = \max\{ \aHhat - \estqlo, \epsilon\},
\end{equation}}for a small $\epsilon = 0.001$.
With this, we calculate the \edit{nonconformity} score as the minimum multiplicative factor for which we get coverage in all dimensions of the output \cite{angelopoulos2021gentle}:
\begin{equation}
\label{eq:nonconformity_score}
    S(\state^{t}, \lowH^{i}, \aH^i) = \inf \{\rho : \aH^t \in (\aHhat^t - \rho \Delta_t^{-}, \hat{a}_t + \rho \edit{\Delta_t^{+})} \}.
\end{equation}

\para{\edit{Calibrating Uncertainty to Target User}}\label{subsec:online_aci_update}
At test time, we receive sequential data, referred to as $\dcalib$, from a target end-user. 
We want to quantify how uncertain is our learned controller when mapping inputs from this target user.
Let $\dcalib = \{(\state^{t}, \lowH^{t}, \aH^t)\}^T_{t=0}$ contain low-dimensional inputs labeled by our target user. 

To perform adaptive conformal quantile regression (ACQR), we initialize our desired miscoverage rate $\alpha_1 = \alpha$ and then seek to compute the empirical quantiles that conformalize the prediction interval. Define time-dependent set $S_{t} = \{S(\state^{i}, \lowH^{i}, \aH^i)\}_{i=1}^t$ as the set of conformal scores for all datapoints in $\dcalib$ up until the current time $t$, where the score function \edit{$S$ itself} is not time-dependent. 

Given a new input $X_{t}$, ACQR constructs a prediction interval for $Y_{t}$ by leveraging the conformity scores obtained via Equation~\eqref{eq:nonconformity_score} on the calibration dataset.  
Finally, for any new human inputs $(\state^{t}, \lowH^{t})$, the calibrated prediction interval is obtained via 
\begin{equation}
C(\state^{t}, \lowH^{t}) = [\aHhat^t - Q_{1-\alpha_t} (S_t) \Delta_t^{-}, \aHhat^t + Q_{1-\alpha_t} (S_t) \Delta_t^{+}) ].
\label{eq:calbirated-interval}
\end{equation} 
where $\lambda_t = Q_{1-\alpha_t} (S_{t}) \coloneqq (1-\alpha_t)(1+1/|\dcalib|)$th empirical quantile of set $S_{t}$. 

To regulate the level of conservativism, at each timestep, we also adjust $\alpha_t$ via the online update
\begin{equation}
    \alpha_{t+1} \coloneqq \alpha_t + \gamma(\alpha - \errt)
    \label{eq:alpha_update}
\end{equation}
\edit{where $\errt$ is defined 
\begin{equation}
  \errt \coloneqq \begin{cases}
    1, & \text{if $Y_t \notin C_t(X_t)$}.\\
    0, & \text{otherwise}.
  \end{cases}
\label{eq:errortime}
\end{equation}
}In practice, we take $\gamma = 0.005$, a value found by \cite{gibbs2021adaptive} to give relatively stable trajectories while being large enough to make meaningful changes to $\alpha_t$.

\para{Coverage Guarantee} 
\edit{Our final guarantees are inherited from ACQR,} which ensures asymptotic $\alpha$ coverage \cite{gibbs2021adaptive}. 

Mathematically, 
\begin{equation}
    \left\vert \frac{1}{T} \sum_{t=1}^T \errt - \alpha   \right\vert \leq \frac{\max\{\alpha_1, 1-\alpha_1\} + \gamma}{T\gamma}
\end{equation}
At $\lim_{T\rightarrow \infty}$, $\lim_{T\rightarrow \infty} \frac{1}{T} \sum_{t=1}^T \errt$ approaches $\alpha$.
This guarantees ACQR gives the $1-\alpha$ long-term empirical coverage frequency regardless of the underlying data generation process. 
\edit{While these asymptotic coverage guarantees are sound in theory, we believe it is important to acknowledge that typical robot deployment conditions are finite-horizon. Nevertheless our empirical findings in Section~\ref{sec:experiments} indicate the practical utility of using ACQR to calibrate black-box assistive controllers compared to naively running the base model.}

\subsection{\edit{Monitoring Uncertainty}}
\label{subsec:monitoring}

With our conformalized teleoperation approach,  the robot can be confident that any user's low-DoF input to $\controller_\theta$ can be associated with an interval of high-DoF actions that is guaranteed to include the human's \textit{desired} high-DoF action (under the assurances provided by ACQR). 
Note that if we had left the regressed quantiles as they are output by $\controller_\theta$ then we have no rigorous notion of uncertainty with respect to the specific end-user. 
In fact, as we see in our experiments in Section~\ref{sec:results}, QR is often overly confident even when it lifts the user's inputs into erroneous high-DoF action spaces. 

With this calibrated notion of uncertainty, we further propose a simple mechanism for detecting high uncertainty in the user inputs seen at test time.  \edit{Recall that our ACQR method yields a high-dimensional (7-DoF) calibrated upper and lower interval bound denoted by $C(\cdot)$ in Equation~\ref{eq:calbirated-interval}. 
We distill the high-dimensional uncertainty into a single scalar measure by considering the radius of the sphere formed by the L2 distance between the upper and lower interval bounds. }
Mathematically, our \textit{scalar uncertainty score} is:
\begin{equation}
    \small 
    U(\state^{t}, \lowH^{t}) \coloneqq ||(\aHhat^t + Q_{1-\alpha_t} (\state^{t}) \Delta_t^{+})) - (\aHhat^t - Q_{1-\alpha_t} (\state^{t}) \Delta_t^{-})||_2.
\label{eq:uncertainty_measure}
\end{equation}
\normalsize
By choosing a threshold $\beta$ of high uncertainty scores, we have a simple switching mechanism which flags high uncertainty at inputs where $U(\state^{t}, \lowH^{t}) > \beta$, and low-uncertainty elsewhere. 
This can be used to stop robot operation, ask for more clarification on the human's intents, or collect more data to improve the model.  
While in this work we do not study exactly \textit{how} this detection mechanism could be used, we do investigate if it can, \edit{with statistical significance}, distinguish between low and high-prediction error states in Section~\ref{sec:results}.

\section{\edit{Evaluation} Setup}
\label{sec:experiments}
To evaluate the efficacy of our proposed approach, we run a series of quantitative and qualitative experiments to study how different training data distributions $\dtrain$ and \edit{user's} inputs ($\dcalib$) impact the learned mapping's uncertainty. 

\para{Types of Uncertainty}

In the context of assistive teleoperation, we study \edit{three sources of real-world uncertainty present in teleoperation data:} \edit{\textit{latent preferences} (at the trajectory level), \textit{low control precision} (at the trajectory level), and \textit{low-dimensional input schemes} (at the input level). }

\smallskip \noindent 
\edit{\textbf{\textit{Trajectory level: Latent preferences.}}} First, we \edit{consider the scenario where the training} distribution $\dtrain$ contains demonstrations from a population of users with varied latent preferences about the task. \edit{We assume all users have the same low-dimensional input scheme (e.g., they all agree on which low-DoF input corresponds to which high-DoF action), but the learning problem is under-specified: a single low-DoF input could correspond to multiple different desired high-DoF actions under two different latent preferences.}
Here, the \edit{\textit{latent preferences}} we consider are which object (i.e., goal) the user wants to move the robot to, and which grasp pose the user wants to pick up a coffee mug with. 

\smallskip \noindent 
\edit{\textbf{\textit{Trajectory level: Low control precision.}}} Second, \edit{we study scenarios where the} training demonstrations exhibit periods of \edit{\textit{low control precision}} or inconsistent task execution. We \edit{assume there is a single goal} but the \edit{users exhibit} noisy behavior on the way towards a goal (such as grasping a cup), \edit{in the training dataset}. \edit{Both training and calibration users use the same low-dimensional input labeling scheme, and the calibration users also exhibit low-precision task execution.}

\smallskip \noindent 
\edit{\textbf{\textit{Input level: Low-dimensional input schemes.}}} \edit{Third, we consider uncertainty that stems from \textit{diverse low-dimensional input schemes}, i.e., the different ways that humans choose low-DoF inputs for the same high-DoF robot trajectory. Here, uncertainty arises because of natural differences between the low-dimensional input schemes seen at training time and the deployment-time users' intuitive understanding of low-dimensional inputs. We obtain calibration data in this scenario by showing the same robot trajectory to three human users, and collecting their low-dimensional input sequence labels.}

\para{Environments}
We study uncertainty in the \edit{\textit{latent preferences} and \textit{low control precision}} contexts through experiments in three controlled environments: in a toy assistive navigation \textbf{GridWorld} environment for building intuition, and a Kinova robotic manipulator \textbf{7DOF goal-reaching} and \textbf{7DOF cup grasping} setting. 
The Gridworld environment is a 25x25 gridworld in which the robot must navigate to achieve a goal state. In the 7DOF settings, expert demonstrators are tasked with either kinesthetically moving the robot towards one of the two objects on the table, or moving the robot to grasp a mug from either the lip or the handle. 
The uncertainty context informs the construction of the training dataset for each domain. 
\edit{We study uncertainty in the \textit{low-dimensional input schemes} context through an experiment in the \textbf{7DOF goal-reaching} setting. To evaluate our method on diverse user input schemes at calibration time, we collected low-dimensional input sequences from a mixture of simulated users and novice human operators (more details in Section \ref{sec:experiments}). }

\para{Baselines}
We compare our method \acqr to vanilla Quantile Regression (\qr), where we train our teleoperation controller but do not calibrate the intervals on the target user online. 
We additionally compare our method to an ensemble uncertainty quantification approach \edit{\ensemble} \cite{lakshminarayanan2017simple}. 
For the \edit{\ensemble} baseline, we train $M=5$ neural networks with the same encoder-decoder structure as in our teleoperation controller design. 
Each model outputs a predicted mean $\mu_{\theta}(\lowH,\state) \in \RR^{n_a}$ and variance $\sigma^2_{\theta}(\lowH,\state) \in \RR^{n_a}$ for the prediction of the high-DoF robot action $a$ intended by the user input $\lowH$ at state $\state$. 
We randomly initialize the model weights and data order. We take the mixture of the multivariate Gaussians as the model prediction, and the first standard deviation from the mean as the prediction interval $C_t(\lowH,\state)$. 

\para{Conformal Hyperparameters}
Our implementation of ACQR uses a step size $\gamma = 0.005$ \cite{gibbs2021adaptive}, target mis-coverage level of $\alpha = 0.1$, and an initial $\alpha_1 = 0.1$. 
Additionally, our proposed detection mechanism (Section \ref{subsec:monitoring}) uses a threshold $\beta$. 
In the \textit{latent preferences} setup, $\beta_{grid} = 1.5$, $\beta_{goal} = 0.05$, and $\beta_{cup} = 0.15$. In the \textit{control precision} context, $\beta_{grid} = 1.0$, $\beta_{goal} = 0.05$, and $\beta_{cup} = 0.12$.

\para{Metrics}
We focus on measures of coverage and prediction interval length. Coverage is defined as 
\begin{equation}
  Cov := \frac{1}{|\dcalib|} \sum_{(\lowH^t,\state^t,\aH^t) \in \dcalib} \mathbbm{1}[\aH^t \edit{\in} C_t(\lowH^t,\state^t)].
\label{eq:coverage_percent}
\end{equation}
We also seek to evaluate the size of the prediction interval, as there exists a tradeoff between coverage and interval sizes. An ideal model has tight intervals but good coverage. 
\edit{For \acqr and \qr, 
interval length for each datapoint is defined as the distance between the upper bound $C_t^{+}$ and lower bound $C_t^{-}$ of the standard deviation of the prediction interval, averaged across each dimension $d \in n_a$:

\begin{equation}
  I_L(\lowH,\state) := \frac{1}{n_a}\sum_{d=1}^{n_a} C_t^{+}(\lowH^t,\state^t)_d - C_t^{-}(\lowH^t,\state^t)_d.
\label{eq:interval_length}
\end{equation}
}
We average final metrics over all datapoints in $\dcalib$.

\section{\edit{Evaluation} Results}
\label{sec:results}

We break down our results into five major takeaways, focusing on
\edit{in-distribution (ID) and out-of-distribution (OOD) calibration users, comparison} of our various uncertainty quantification methods, \edit{and the performance of} our proposed detection mechanism. 

\medskip 
\noindent \textit{\textbf{Takeaway 1:} Even when \edit{users operate with an in-distribution input scheme on in-distribution high-DoF trajectories}, an uncalibrated mapping $\controller_\theta$ (\qr) miscovers the human's desired high-DoF action more than \acqr.}
\medskip 

\begin{figure}[t]
    \centering
    \includegraphics[width = 0.49\textwidth]{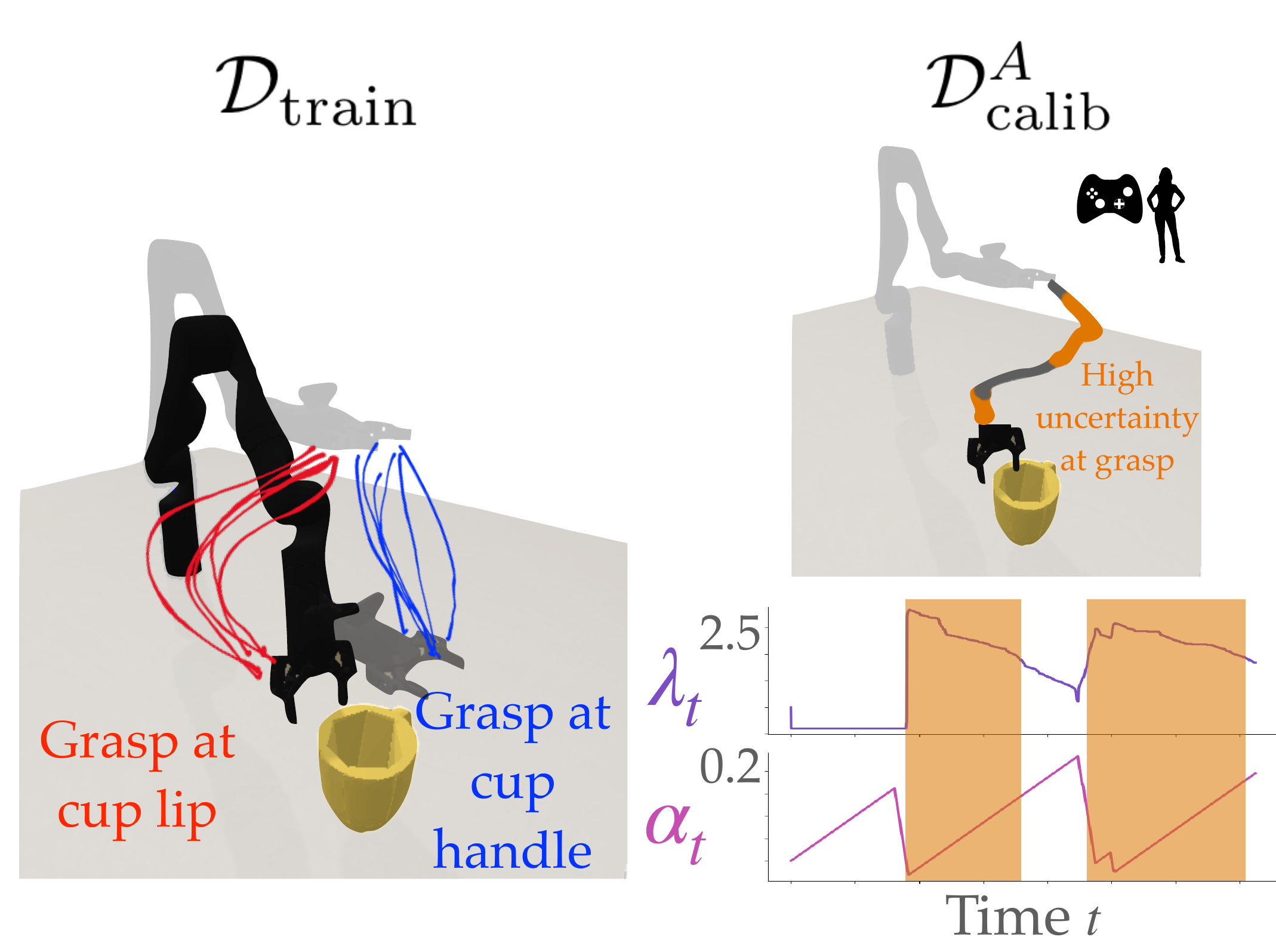}
    \caption{\textbf{7DOF cup-grasping with latent preferences.} (left) $\dtrain$ exhibits multimodality in grasp. 
    (right) When ACQR calibrates to $\dcalib^A$. Orange dots indicate timesteps with high uncertainty ($U > \beta_{cup}$). $\lambda_t$ represents the multiplicative factor by which the quantile intervals are expanded. 
    While $\alpha_t$ increases initially, it decreases later as the robot orients itself towards the cup and as it approaches the cup, and uncertainty in the desired high-dimensional action increases. 
    Correspondingly, $\lambda_t$ increases at the points of uncertainty. 
    The orange highlights denote timesteps where the uncertainty \edit{is greater than threshold} $\beta_{cup}$ and aligns temporally with the orange points on the trajectory.
    For grasps on the lip, there is higher uncertainty at the start of the trajectory and at the grasp location, due to the variance in lip grasp demonstrations in the training dataset. }
    \label{fig:preference_context_cup_grasping}
\end{figure}

We highlight this takeaway in the setting of \textbf{7DOF Cup-grasping} with diverse \textbf{latent preferences}, but further results can be found in the supplementary. 
Recall that in this setting the demonstrators may pick up a cup from the handle, others from the lip. Thus, $\dtrain$ consists of 14 expert demonstration trajectories, where half pick up the cup from the handle, and half pick up the cup from the lip (shown in left of Figure \ref{fig:preference_context_cup_grasping}). 
We calibrate $\controller_\theta$ on an unseen user, referred to as Alice and denoted $\dcalib^A$, who gives 3 demonstrations of picking up the cup from the lip. \edit{In this case, the calibration demonstrations of target user, Alice, were provided by one of the researchers who also provided expert demonstrations in the training data.}
Both the demonstrators and target user employ a heuristic strategy to deterministically annotate $\lowH^t$ for consecutive state pairs, where $\lowH^t$ is the change in $x$-direction and $y$-direction of the end effector from $\state^t$ to $\state^{t+1}$. 
We calibrate to each held-out trajectory, simulating the inputs to the assistive teleoperation controller over time as though the user was controlling the robot in real time. 

Using \acqr, we see that uncertainty is highest at the \textit{start} and \textit{end} of the interaction when the human has to give the final inputs to orient the robot arm to face downward to achieve their desired cup grasp (shown in right, Figure \ref{fig:preference_context_cup_grasping}). 
Intuitively, $\dtrain$ contains higher disagreement amongst the training data generators as they position the robot for grasping.
These critical states \cite{huang2018establishing} are informed by the specific user's preferences. Without additional context and due to the underspecified input, the robot cannot be certain about the correct way to map the user's low-DoF input to a high-DoF action. 
Quantitatively, \qr achieves $52.7\%$ coverage on $\dcalib^A$, while \acqr achieves $92.6\%$ coverage (where target coverage is $90\%)$. 
This result demonstrates that even for an end-user providing in-distribution demonstrations, adaptively calibrating to unseen data is necessary for achieving informative uncertainty bounds. 

\begin{figure}[t!]
    \centering
    \includegraphics[width = 0.49\textwidth]{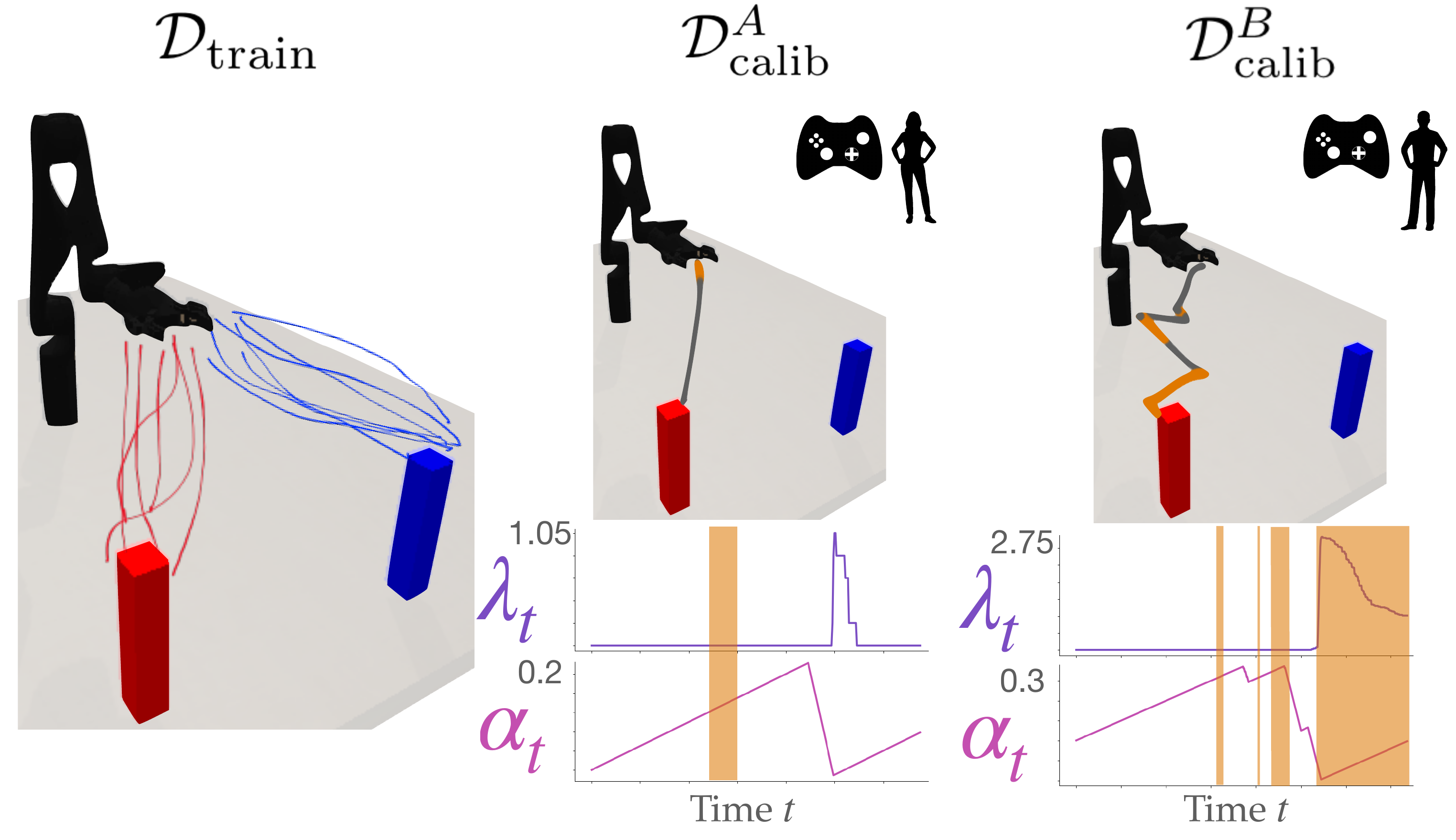}
    \caption{\textbf{7DOF goal-reaching with latent preferences.} (left) $\dtrain$ contains teleoperation trajectories towards red and blue goals. (center) Alice's demonstration contains high uncertainty only at the beginning of the task. (right) Bob's indirect path flags high uncertainty throughout. ACQR maintains lower $\lambda_t$ and higher $\alpha_t$ throughout $\dcalib^A$ demonstration than throughout $\dcalib^B$.}
    \label{fig:preference_context_two_goals}
\end{figure}

\medskip 
\noindent \textit{\textbf{Takeaway 2:} \edit{When users provide in-distribution low-dimensional inputs on out-of-distribution calibration trajectories}, \acqr can expand uncertainty when necessary but also contract it for inputs that align with its training distribution.}
\medskip 

We highlight this takeaway in the setting of \textbf{7DOF Goal-reaching} with diverse \textbf{latent preferences}.
Here, $\dtrain$ consists of 120 expert demonstrated trajectories, where half of the demonstrators prefer the blue goal and half prefer the red goal (see left, Figure \ref{fig:preference_context_two_goals}). 
As in the cup-grasping domain, demonstrators and target users employ a heuristic strategy to deterministically generate $\lowH^t$ for consecutive state pairs, using the change in $x$-direction and $y$-direction of the end effector. 
We calibrate on two target user profiles, Alice and Bob. Alice represents an in-distribution user: she demonstrates 6 near-expert trajectories to each of the two goals, taking direct paths to the objects. Her data is denoted $\dcalib^A$.
The second user, Bob, takes 6 extremely indirect paths towards each goal, representing our out-of-distribution user; his data is denoted $\dcalib^B$. \edit{In this case, calibration data for Alice and Bob was provided by a researcher who teleoperated in accordance with the particular user profile.}

In the center and right of Figure~\ref{fig:preference_context_two_goals}, we see qualitatively that \acqr run on demonstrations in $\dcalib^A$ contain fewer instances of high uncertainty, only flagging the start states as uncertain since the assistive mapping isn't confident \edit{in} which \edit{initial high-DOF actions} the human wants to \edit{take}. However, input data from $\dcalib^B$ are frequently flagged as high uncertainty. Interestingly, there are several key portions of the trajectory wherein even the out-of-distribution user appear to be well-predicted by the model. 

Quantitatively on $\dcalib^A$, \qr achieves only $82.1\%$ coverage on $\dcalib^A$ while \acqr achieves $91.9\%$ coverage. More surpisingly, on the out-of-distribution user from $\dcalib^B$, \qr is overly confident, achieving only $68.8\%$ coverage. Instead, \acqr achieves $90.2\%$ coverage. 
This result demonstrates that $\controller_\theta$ can confidently mispredict the intended actions of an out-of-distribution user, but ACQR can adjust the prediction intervals to handle such interaction data.

\begin{figure}[tb]
    \centering
    \includegraphics[width = 0.5\textwidth]{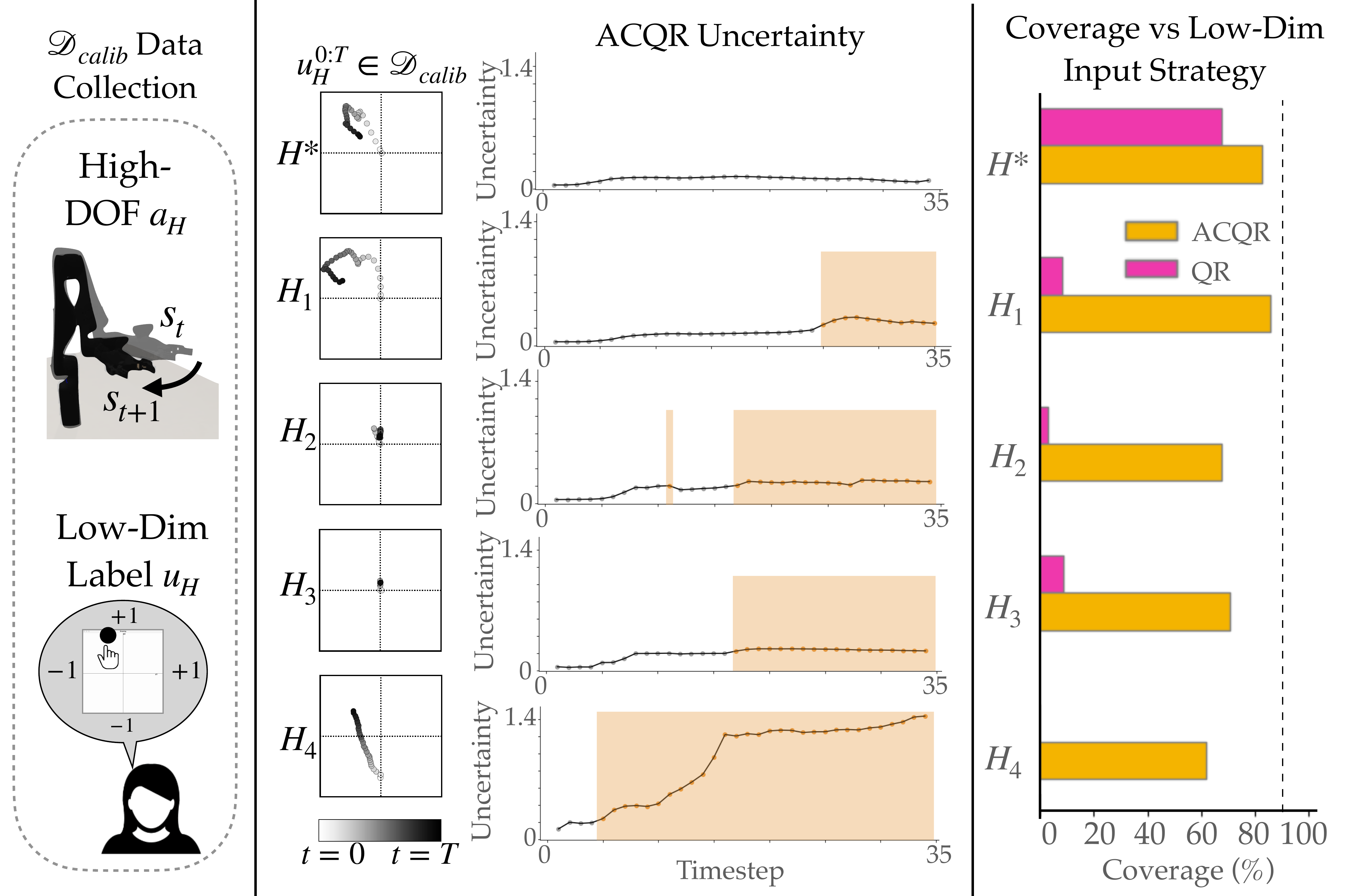}
    \caption{\edit{\textbf{7DOF goal-reaching (blue) with out-of-distribution low-dimensional input schemes.} (Left) The calibration data is collected by querying a mixture of simulated and human users for their desired low-dimensional input for a given high-DOF robot action.} \edit{(Middle) Our uncertainty monitoring mechanism, at the threshold $\beta_{goal,human}=0.2$, flags high uncertainty for out-of-distribution input schemes $H_1-H_4$.} \edit{(Right) Calibration using \acqr increases coverage over \qr on the calibration data from all in- and out-of-distribution users, increasing coverage towards, but not yet at, the desired level of 90\%.}}
    \label{fig:blue_coverage}
\end{figure}

\medskip 
\noindent \edit{\textit{\textbf{Takeaway 3:} As users' low-dimensional input schemes become more out-of-distribution, \acqr adapts to cover their desired high-DoF actions more than \qr.}}
\medskip  

\edit{
We highlight this takeaway in the task of \textbf{7DOF Goal-reaching}. We use the base model trained from the \textbf{latent preferences} context. Training demonstrations are labeled with low-dimensional inputs via a deterministic, heuristic strategy using the change in $x$-direction and $y$-direction of the end effector. 

In the calibration phase, target users provide low-dimensional input labels for a single calibration trajectory to the blue goal (see Section  \ref{subsec:red_goal_low_dim_variance} for results on a trajectory to the red goal). 
For each trajectory, we collected the low-dimensional input labels for pairs of consecutive robot states, downsampled along the full trajectories. We considered five input schemes: three real human annotators ($H_2, H_3, H_4$), one simulated user operating under an alternative heuristic scheme ($H_1$), and one in-distribution simulated user ($H^*$). 

Specifically, $H_1$ gives input $\lowH$ equal to the change in $z$-direction and $y$-direction of the end effector, representing a slightly out-of-distribution low-dimensional input scheme. Users $H_2, H_3, H_4$ are three real, human users with technical expertise but novice to the task of robot teleoperation with low-dimensional control (see left hand side of Figure \ref{fig:blue_coverage} for the query setup).

In Figure \ref{fig:blue_coverage} (right), we see that the base model without calibration, \qr, achieves 68\% percent coverage on the in-distribution user $H^*$, but achieves poor coverage ($<10\%$) on out-of-distribution input schemes ($H_1$-$H_4$). 
In contrast, calibration using \acqr increases coverage on the calibration data from all in- and out-of-distribution users, increasing coverage towards the desired level of 90\%. Note that because \acqr gives a coverage guarantee asymptotic in time, the coverage achieved by \acqr nears but does not reach the desired level given the limited trajectory length. 

Next, we study how our uncertainty monitoring mechanism is influenced by OOD input schemes. In Figure \ref{fig:blue_coverage} (middle), our uncertainty monitoring mechanism (at the threshold $\beta_{goal,human}=0.2$, increased from 0.05 for noisy human inputs) flags high uncertainty more frequently as the user input schemes become more OOD. 
For example, user $H_1$, is relatively in-distribution: they give input via the $z$- and $y$-displacement of the end effector which is similar to user $H^*$ from the training population. 

Because of the similarity in input sequence, \acqr's uncertainty intervals are not as large for $H_1$ and thus the monitor only activates briefly along the middle of the trajectory (for $12$ timesteps). 
On the other extreme, for our most OOD user, $H_4$, our monitor activates directly after the first input and continues to flag each consecutive input as uncertain. Finally, for users $H_2$ and $H_3$ whose input schemes are extremely similar in pattern but still quite OOD, our uncertainty monitor shows a similar detection pattern, indicating that our monitor's behavior is consistent. 

}

\medskip 
\noindent \textit{\textbf{Takeaway 4:} While ensembles can sometimes capture the high-uncertainty of the learned mapping, they are highly sensitive to the problem domain and in general do not come with any assurances.}
\medskip 

\begin{figure}[t]
    \centering
    \includegraphics[width = 0.49\textwidth]{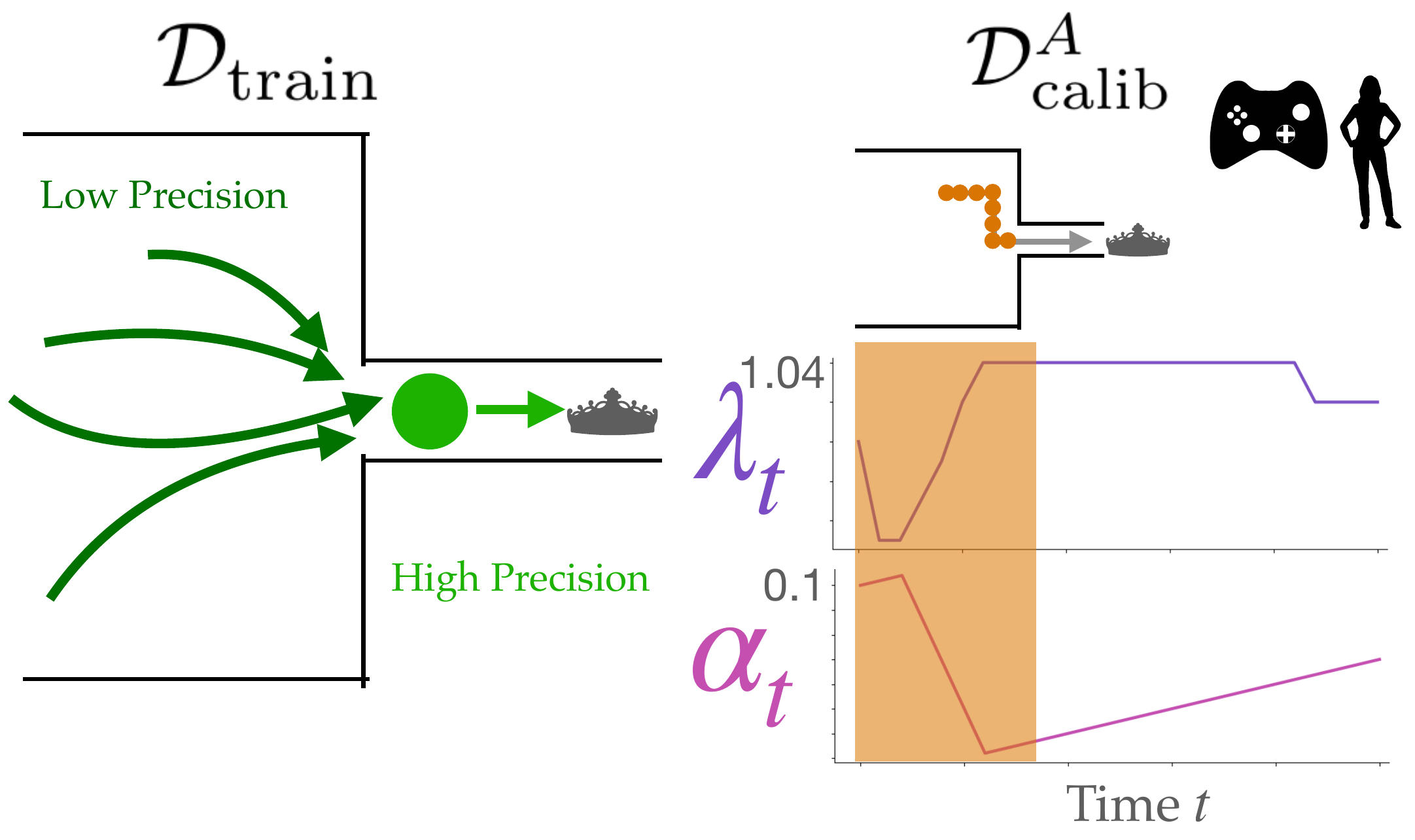}
    \caption{\textbf{GridWorld with low control precision}. The areas of highest uncertainty ($\beta = 1.0$) occur as the heldout demonstration approaches the tunnel. Inside the tunnel, uncertainty is low. $\alpha_t$ decreases throughout the approach to the tunnel, and increases once inside the tunnel.}
    \label{precision_grid}
\end{figure}

We highlight this takeaway in the setting of \textbf{GridWorld} and \textbf{7DOF Cup-grasping} settings under uncertainty induced by \textbf{low-control precision}. 
In the low-control precision context, our target user is the same as the demonstrator. 
The GridWorld domain requires demonstrators to navigate from an open region through a narrow tunnel to reach the goal state (Figure \ref{precision_grid}). 
Trajectories in $\dtrain$ begin at random locations in the open region with noisy actions \edit{representing variance in paths taken at less critical regions}. 
Once the user is in the hallway, \edit{a narrow region where precision is critical}, there is no stochasticity in their actions. For calibration, $\dcalib^A$ contains 36 heldout trajectories of the user moving to the goal from different starting states. 
Intuitively, we find that the areas of highest uncertainty occur at the low-precision parts of the task, where there are, for the target user, many paths to approach the hallway. 
Once inside the tunnel, uncertainty is low (right, Figure \ref{precision_grid}). In this domain, we see \acqr achieve 92.7\% coverage with mean interval lengths of 0.86 while the \ensemble struggles to achieve the target coverage (81.7\%) with mean interval lengths (0.697). 

Although \ensemble mispredicts the actions more frequently than \acqr in \textbf{Gridworld} setting, we see different results in \textbf{7DOF Cup-Grasping} environment (see Table \ref{tab:baselines_context_2_A}). Here, $\dtrain$ contains 7 kinesthetic demonstrations by a single user where all demonstrations pick up the cup by the handle. 
The trajectories are very precise near the handle, but less precise in the path towards the cup where there \edit{is} more free space. 
We calibrate on the same target user, $\dcalib^A$, who demonstrates 3 unseen test trajectories of picking up the cup by the handle. 
Empirically, \ensemble exhibits higher coverage than ACQR: 
\acqr achieves 93.8\% coverage with mean interval lengths of 0.044, while \ensemble achieves 100\% coverage with a larger mean interval length of 0.745. 
While empirically our ACQR algorithm achieves coverage of $1-\alpha = 0.9$ in both environments, the Ensemble does not provide any statistical guarantees and coverage can vary significantly between environments. 
We hypothesize that the latter point stems from the training action distribution: in the Gridworld, the training action distribution is multi-modal around the values $\{-1,0,1\}$, inducing high variance; in the 7DOF robot domain, the robot’s joint velocities are bounded to $[-1,1]$ per-joint and the training action distribution has small variance centered around the mean, making mis-prediction less likely.

\begin{table}[tbp!]
\centering
{
\caption{Low-Control Precision: Coverage and Interval Size for $\dcalib^A$}
\label{tab:baselines_context_2_A}
\renewcommand{\arraystretch}{1.5}
\begin{tabular}{c c  c  c  c  c  c} 
\toprule 
 \multicolumn{1}{c}{\textbf{Algorithm}} &  \multicolumn{3}{c}{\textbf{Coverage}} & \multicolumn{3}{c}{\textbf{Interval Length}}\\
\hline 
{} &  Grid & Goal & Cup &  Grid & Goal & Cup \\
{} &  $\dcalib^A$ & $\dcalib^A$ & $\dcalib^A$ & $\dcalib^A$ & $\dcalib^A$ & $\dcalib^A$ \\
 [0.5ex] 
 \hline
\acqr  &  0.927 & 0.890   & 0.938  & 0.860 & 0.009 & 0.044 \\
\ensemble   &  0.817 & 1.0  & 1.0  & 0.697 & 0.745  & 0.742 \\
\qr   &  0.910 &  0.994  & 0.402  & 0.689 & 0.012 & 0.035 \\
 \bottomrule
\end{tabular}
}

\end{table}

\begin{figure}[t]
    \centering
    \includegraphics[width = 0.49\textwidth]{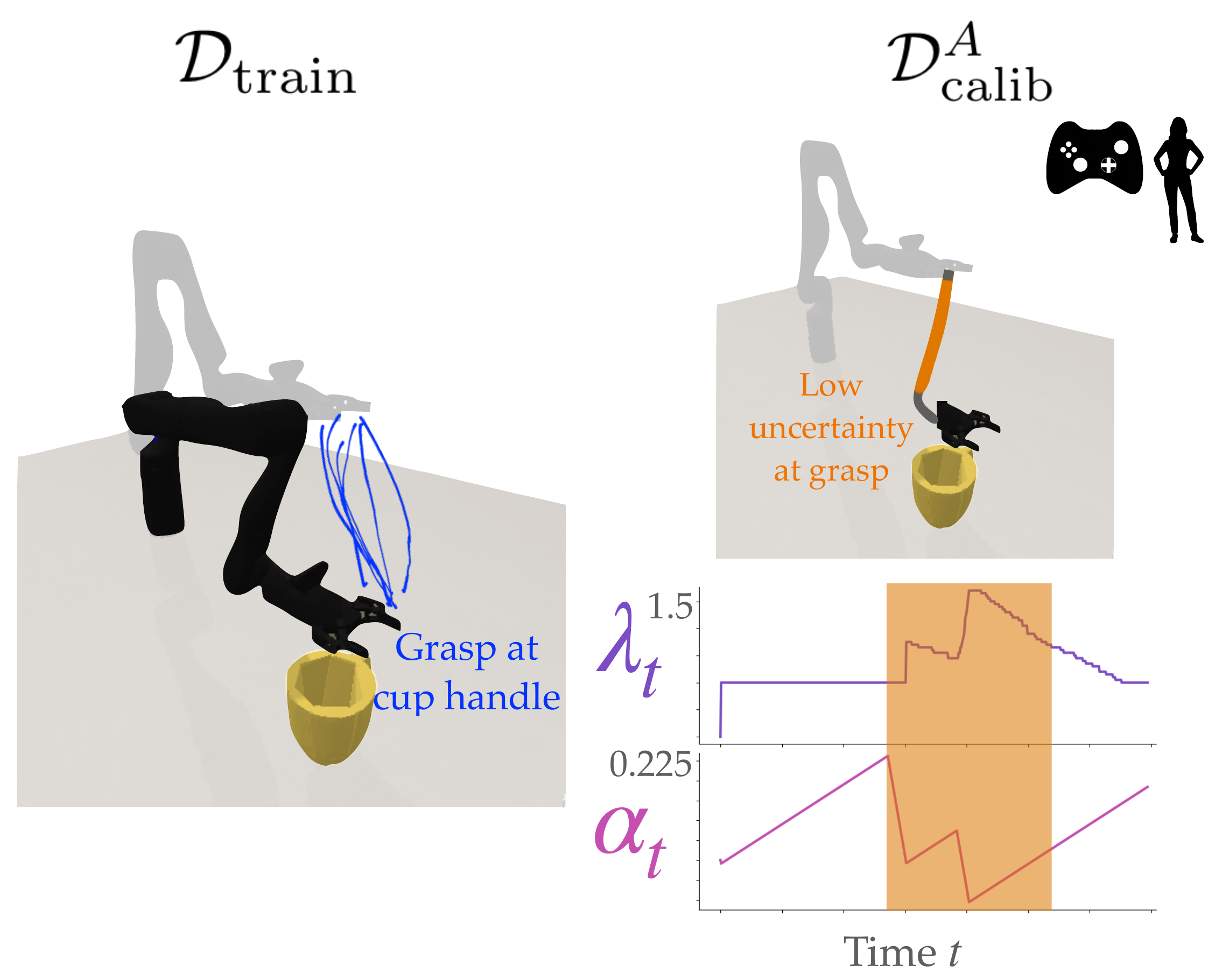}
    \caption{\textbf{7DOF Cup-grasping with low control precision.} Demonstrations from a single user position the gripper to grasp a cup at the handle (left). ACQR flags high uncertainty in the path towards the cup, but has the lowest uncertainty at the grasp location (right), where high precision is required.}
    \label{cup_train_on_one}
\end{figure}

\begin{table*}[t!]
\centering
{
\caption{Prediction Error in Contexts with Uncertainty Induced by Low-Precision}
\label{tab:prediction_error_context2}
\renewcommand{\arraystretch}{1.5}
\begin{tabular}{c  c  c | c  c  c} 
 \toprule 
 \textbf{Environment} & \textbf{Target User} & \begin{tabular}{@{}c@{}} $\beta$ \end{tabular} & \begin{tabular}{@{}c@{}} $\aHhat$ \textbf{Error at Uncertain Inputs}: \edit{mean, (std)} \end{tabular} & \begin{tabular}{@{}c@{}} $\aHhat$ \textbf{Error at Certain Inputs} \end{tabular} & \textbf{T-test} \\ 
 [0.5ex] 
 \hline
 GridWorld & $\DD^A_{calib}$ & 1.0 & 0.680, (0.415) & 0.040, (0.073) & \begin{tabular}{@{}c@{}}$t(966)=36.62$  $p<0.001$  \end{tabular} \\
 7DOF Goal & $\DD^A_{calib}$ & 0.05 & 0.014, (0.005) & 0.006, (0.001) & \begin{tabular}{@{}c@{}}$t(3109)=55.47$  $p<0.001$  \end{tabular} \\
 7DOF Cup & $\DD^A_{calib}$ & 0.12 & 0.018, (0.008) & 0.010 (0.002) & \begin{tabular}{@{}c@{}}$t(2041)=36.57$ $p<0.001$  \end{tabular} \\
 \bottomrule
\end{tabular}
}
\end{table*}

\medskip 
\noindent \textit{\textbf{Takeaway 5:} Our proposed detection mechanism based on the uncertainty interval size is an effective way to separate high and low uncertainty states and inputs.}
\medskip  

Finally, we seek to verify the hypothesis that instances of high uncertainty in the user's demonstration at test time can be meaningfully identified through our proposed detection mechanism.

To test this, for each environment, we choose an uncertainty threshold, $\beta$ (tuned per  domain) and test if there is a statistically significant difference in prediction error $||\hat{a}_t - a_t||_2$ at states with high uncertainty (uncertainty above $\beta$) and states with low uncertainty (uncertainty below or equal to $\beta$). 
While ultimately this is a design parameter, we aim to understand if even simple choices of $\beta$ demonstrate statistically meaningful separation between the classification of certain and uncertain states. 

If this is true, then this indicates the potential for using our approach to proactively flag potential robot low-to-high dimensional mapping errors before they occur and ask for further assistance from the human. 

We detail the results in the \textbf{7DOF Goal-reaching} domain with \textbf{low-control precision} uncertainty. 
Specifically, we perform a T-test of unequal variances \cite{kim2015t} between the distribution of prediction errors for inputs flagged as anomalous by our mechanism and the distribution of prediction errors for inputs flagged as nominal on $\dcalib^A$ (see Table \ref{tab:prediction_error_context2}).
We find that our mechanism separates low and high uncertainty inputs in a statistically significant manner ($p < 0.001$ in this environment). \edit{We evaluate distributional separation using a T-test to confirm the prediction errors from low and high uncertainty inputs come from different distributions.} Furthermore, the mean prediction error is higher in uncertain states than certain states. 
We find this result particularly promising, because this indicates that \acqr can be used as a principled detection mechanism for mitigating the robot's learned mapping errors and potentially providing a way of proactively asking the user for assistance.

\section{\edit{Discussion \& Conclusion}}
\label{sec:conclusion}
\para{Discussion} 
One limitation of our approach is that the model cannot differentiate between these different types of semantic uncertainty (e.g., preferences vs. control precision). 
Here, we are excited to investigate how additional context (e.g., images \cite{karamcheti2021learning} or language \cite{karamcheti2022lila}) can reduce the robot's uncertainty intervals. 

Future work should also investigate seeking appropriate assistance from human users, through other modalities, like language \cite{jiang2021can, ren2023robots, karamcheti2022shared}.

\para{Conclusion}
In this work, we present Conformalized Teleoperation, an approach for quantifying uncertainty in learned assistive teleoperation controllers that map from low-DoF human inputs to high-DoF robot actions. 
We find that the dataset with which the controller was trained on significantly impacts the robot's ability to infer the user's intended high-DoF action, and off-the-shelf controllers can be over-confident even when interacting with significantly out-of-distribution end-users. However, we find that our proposed adaptive conformal quantile regression approach can meaningfully detect these state and inputs that cause uncertainty, providing a promising path forward towards the detection and mitigation of such assistive teleoperation failures.

\section*{Acknowledgments}
We’d like to thank Anastasios Angelopoulos, Ravi Pandya, Pat Callaghan, and Suresh Kumaar Jayaraman, whose feedback and comments helped strengthen the paper. Finally, MZ’s PhD research is generously supported by the NDSEG fellowship.

\bibliographystyle{plainnat}
\bibliography{rss_references}

\clearpage
\section{Appendix}
\label{sec:appendix}

\subsection{Background: Adaptive Conformalized Quantile Regression (ACQR)} 
\label{sec:cqr_background}

Conformal prediction (CP) is a technique for constructing prediction sets on the output of a model that are guaranteed to contain the target output with specified high probability (i.e., coverage) \cite{vovk2005algorithmic}. 

While there are many variants of CP (see \cite{angelopoulos2021gentle} for an overview), we focus on Adaptive Conformalized Quantile Regression (ACQR) \cite{romano2019conformalized}, an instance of Adaptive Conformal Inference \cite{gibbs2021adaptive}. We ground our work in this technique since it is relevant to regression problems, yields asymptotic coverage guarantees without strong distributional assumptions, and gives tight uncertainty bounds at each input \cite{gibbs2021adaptive, romano2019conformalized}. 

\para{Setup: Quantile Regression (QR)} Let $\dtrain = \{(X_t, Y_t)\}^n_{t=1}$ be a training dataset of $n$ samples which are drawn from an arbitrary joint underlying distribution, $P_{XY}$. 
We seek to \edit{train a regression} model $f$ which takes as input a data point $X$ and outputs $\hat{Y}$, and is trained on the data $\dtrain$. 

In addition to a point prediction ($\hat{Y}_t)$, the regressor is modified to output the \textit{estimated upper and lower conditional quantiles} during training:
\begin{equation}
     \{\estqlo(X_t), \hat{Y}_t, \estqhi(X_t)\} \leftarrow f(X_t), ~~\forall (X_t, Y_t) \in \dtrain,
\end{equation}
where $\estqlo(X_t)$ is an estimate of the $\alpha_{lo}$-th conditional quantile and $\estqhi(X_t)$ is the $\alpha_{hi}$-th quantile estimate. During training, $\estqlo(X_t)$ is learned with an additional Pinball loss \cite{koenker1978regression, romano2019conformalized}. 
Unfortunately, the resulting empirical conditional prediction interval $\hat{C}(X_t) = [\estqlo(X_t), \estqhi(X_t)]$ is not necessarily calibrated. 

\para{Adaptive Conformal Quantile Regression (ACQR)} 

Assume the datapoints we see at test time are not i.i.d, and the distribution generating the data is non-stationary. 
We want to adaptively calibrate a parameter $\alpha_t$ which will continually adjust the estimated prediction intervals $\hat{C}_t(X_t)$, yielding time-dependent intervals denoted by the subscript $t$. Here, we assume access to data online, $\dcalib$, to adapt the prediction intervals based on past performance of the regressor using a variant of adaptive conformal prediction  \cite{gibbs2021adaptive}.

The critical step in ACQR is to, at each time $t$, compute the \textit{conformity score}, a quantification of error on the calibration set. $\dcalib$ is a heldout sequence $\{(X_t, Y_t)\}^T_{t=1}$ which we see online. At each time $t$, we observe the new datapoint in $(X_t, Y_t) \in \dcalib$, and compute the conformity score $S$ as:
\begin{equation}
    S(X_t, Y_t; f) = \max\{\estqlo(X_t) - Y_t, Y_t - \estqhi(X_t)\},
    \label{eq:conformity-score}
\end{equation}
which is the coverage error induced by the regressor's quantile estimates. 

Let $S_t$ be the set of conformity scores for all data points through time $t$ in $\dcalib$. 
In general, the magnitude of $S(X_t, Y_t; f)$ is determined by the miscoverage error and its sign is determined by if the true value of $Y_t$ lies outside or inside the estimated interval. This quantity enables us to conformalize the predicted quantiles and appropriately adjust the estimated interval to account for over- and under-coverage.

\para{Empirical Coverage Online Update}
Our goal in conformal prediction is to, for some miscoverage rate $\alpha$, construct prediction intervals such that the probability that the prediction interval contains the correct label is almost exactly $1-\alpha$. 

Given a new input $X_{t}$, ACQR constructs a prediction interval for $Y_{t}$ by leveraging the conformity scores obtained via Equation~\eqref{eq:conformity-score} on the calibration dataset.  
Mathematically, the calibrated prediction interval for $Y_{t}$ is:
\begin{align}
    C_t(X_{t}) = \big[&\estqlo(X_{t}) - Q_{1-\alpha_t}(S_t), \\~&\estqhi(X_{t}) + Q_{1-\alpha_t}(S_t) \big], 
\end{align}
where the $Q_{1-\alpha_t}(S_t) := (1-\alpha_t)(1+\frac{1}{|\dcalib|})$-th adaptive empirical quantile of $S_t = S_{t-1} \cup S(X_t, Y_t; f)$ conformalizes the prediction interval. Throughout this manuscript, we will use the shorthand $\lambda_t := Q_{1-\alpha_t}(S_t)$ to refer to the adaptive empirical quantile.
Given the non-stationarity of the data distribution, we examine the empirical miscoverage frequency of the previous interval, and then decrease or increase a time-dependent $\alpha_t$, which will asymptotically provide $1-\alpha$ coverage \cite{gibbs2021adaptive} (see next section on coverage guarantee of ACQR). \edit{With $\errt$ defined as in Equation \ref{eq:errortime}, and fixing step size parameter $\gamma > 0$, we perform the online update $\alpha_{t+1} \coloneqq \alpha_t + \gamma(\alpha - \errt)$ (Equation \ref{eq:alpha_update}).}

\subsection{Visualizing Calibrated Intervals}
\begin{figure}[t]
    \centering
    \includegraphics[width=0.5\textwidth]{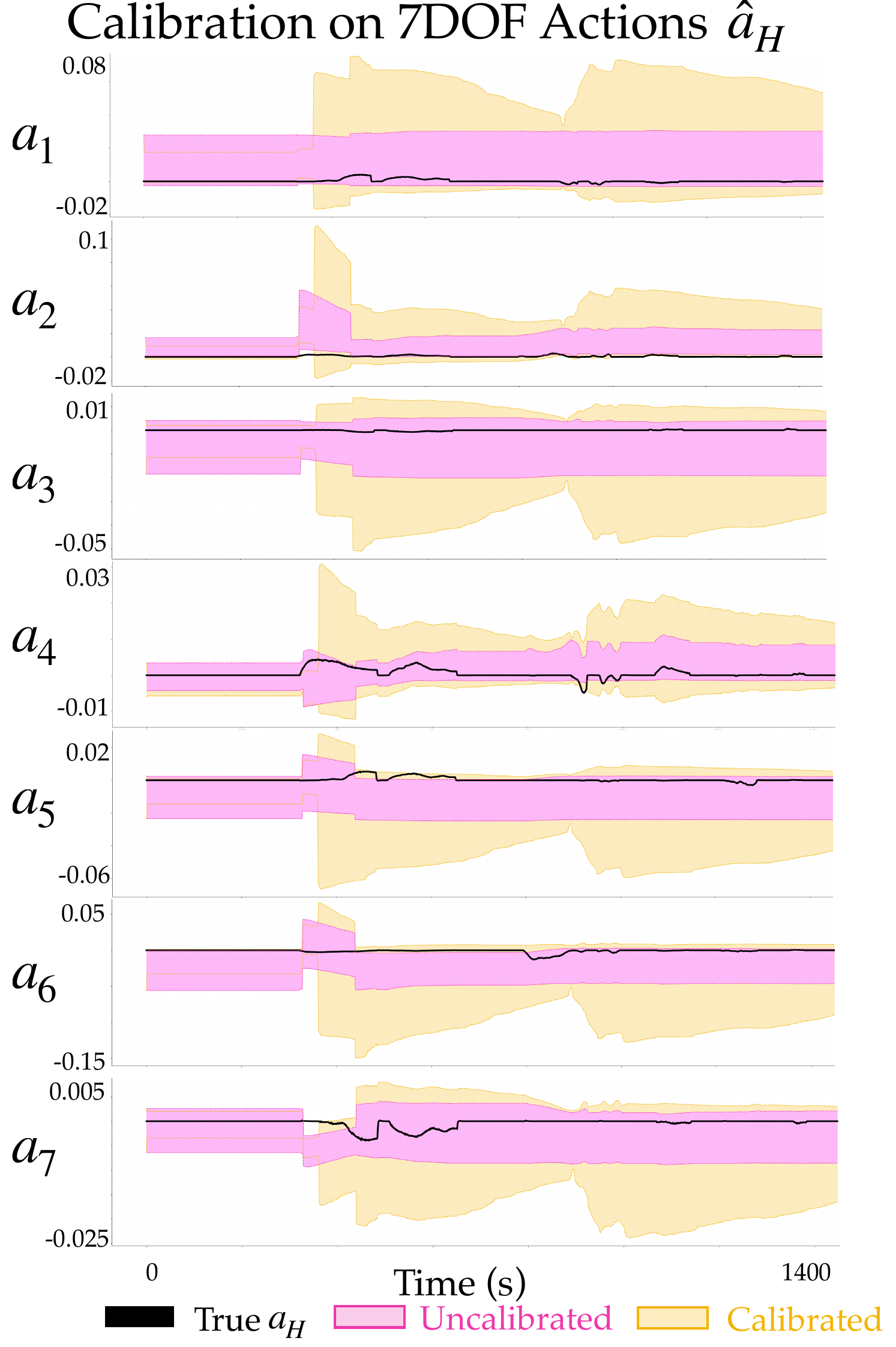}
    \caption{While the uncalibrated interval (pink) mispredicts the true value of the joint angle (black), calibration (orange) expands the size of the intervals such that the mispredicted values become covered.}
    \label{fig:calibrated_intervals}
\end{figure}

Figure \ref{fig:calibrated_intervals} provides a visual of the uncalibrated and calibrated intervals for the joint angles of a trajectory performing the \textbf{7DOF Cup-Grasping} task. The true values of the wrist joint angle for the selected trajectory are shown in black. As the uncalibrated interval (pink) mispredicts, calibration (orange) expands the size of the intervals such that the mispredicted values become covered.

\subsection{Details on Experimental Setup}
This section provides details on environments which are not highlighted in the results section of the paper.

\subsubsection{Latent Preferences}
The \textbf{latent preferences} context examines ACQR behavior in scenarios in which the variance in the intended high-dimensional action stems from demonstrations comprising the training dataset by users with varied behaviors or preferences.

\begin{figure}[t]
    \centering
    \includegraphics[width = 0.49\textwidth]{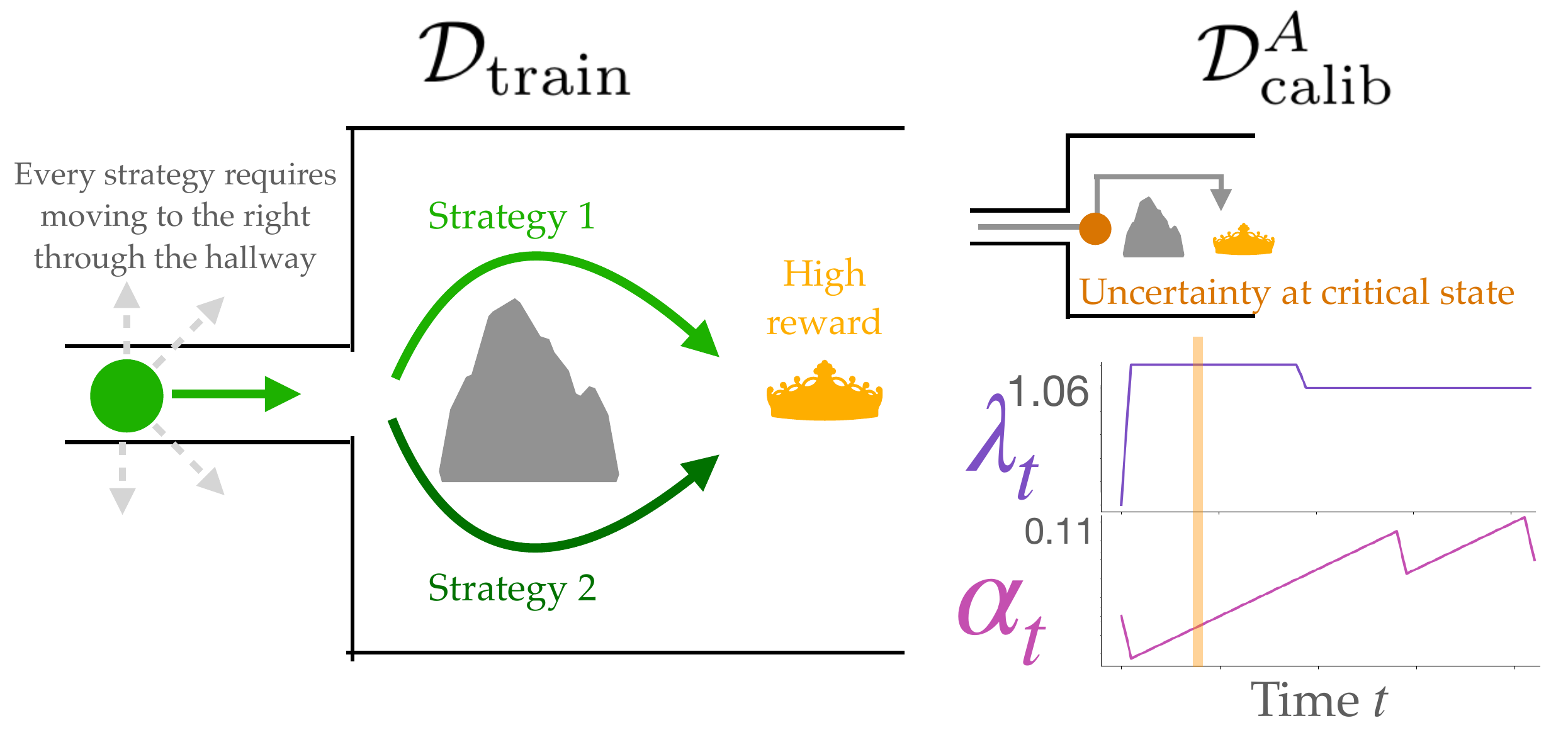}
    \caption{The two preferences present in $\dtrain$ for the gridworld task are moving to the left and right of the obstacle (left). We show a stylized trajectory from target user, Alice ($\dcalib^A$). For $\dcalib^A$, ACQR flags uncertainty above a threshold of 1.5 at the critical state where the two preferences branch, indicated by the orange dot. $\alpha_t$ decreases at the point of uncertainty, and increases for the rest of the trajectory, as the model continuously predicts the human actions correctly. $\lambda_t$ represents the multiplicative factor by which the quantile intervals are expanded. $\lambda_t$ increases at the point of uncertainty, as $\alpha_t$ decreases, but rises as the model gains confidence once the agent has passed the critical state. The orange bar reinforces timesteps where the uncertainty was above the defined threshold. }
    \label{fig:pref_grid}
\end{figure}

\para{\textbf{Task 1: GridWorld}}
We constructed this Gridworld domain as an oversimplified environment to demonstrate the context of latent preferences (Figure \ref{fig:pref_grid}). The task is a 25x25 gridworld in which the demonstrating agent must navigate through a narrow hallway and then around a large obstacle to achieve the goal state (Figure \ref{fig:pref_grid}). $\dtrain$ contains 48 noiseless trajectories, where half move up and around the obstacle, and the other half move down and around the obstacle. The state $s$ is the position of the agent, and the high-DoF action $a$ is the movement in the $x$-direction and $y$-direction of the agent. The low-DoF human action $h$ is $h=1$ for all timesteps, representing the most salient direction of rightward movement throughout the full trajectory. Because of this underspecified input at the critical state where the two strategies branch, the prediction interval constructed by the model should be large. 

We want to verify that ACQR correctly identifies states with high uncertainty at the critical state. To do so, we calibrate on a user, Alice, denoted $\dcalib^A$, who demonstrates 18 trajectories of moving up and around. We find that our detection mechanism with threshold at $\beta=1.5$ correctly identifies uncertainty at the critical state where the two preference modes diverge (Figure \ref{fig:pref_grid}). This tells us that the controller may not correctly produce Alice's desired high-DoF action at the critical state, and the critical state is one where the robot should ask for additional context or intervention.

\subsubsection{Low-Control Precision}
The \textbf{low-control precision} context examines ACQR behavior in scenarios in which the variance in the intended high-dimensional action occurs when the training distribution $\dtrain$ contains inconsistent demonstrations from a single user. 

\edit{
\subsubsection{Low-Dimensional Input Schemes}
The \textbf{low-dimensional input schemes} context examines ACQR behavior in scenarios in which the variance exists not in the intended high-dimensional robot actions, but in the different ways in which human operators give low-dimensional inputs to execute the same high-DOF action. We examine this type of uncertainty as being present only in the calibration dataset. We collected data from three, real human operators, providing labels on pairs of robot states throughout two calibration trajectories. The calibration trajectory to the blue goal contained 35 timesteps, and the trajectory to the red goal contained 100 timesteps. Human operators labeled the robot action taken at each timestep. For each query, operators were able to replay the robot action as many times as needed to determine a low-dimensional input label. Operators provided the label through a graphical interface, where they clicked on a screen to provide what would represent their precise joystick input (see Figure \ref{fig:blue_coverage} (left) to understand the query setup). 
}

\para{\textbf{Task 2: 7DOF Goal-Reaching}}
\begin{figure}[tb]
    \centering
    \includegraphics[width = 0.49\textwidth]{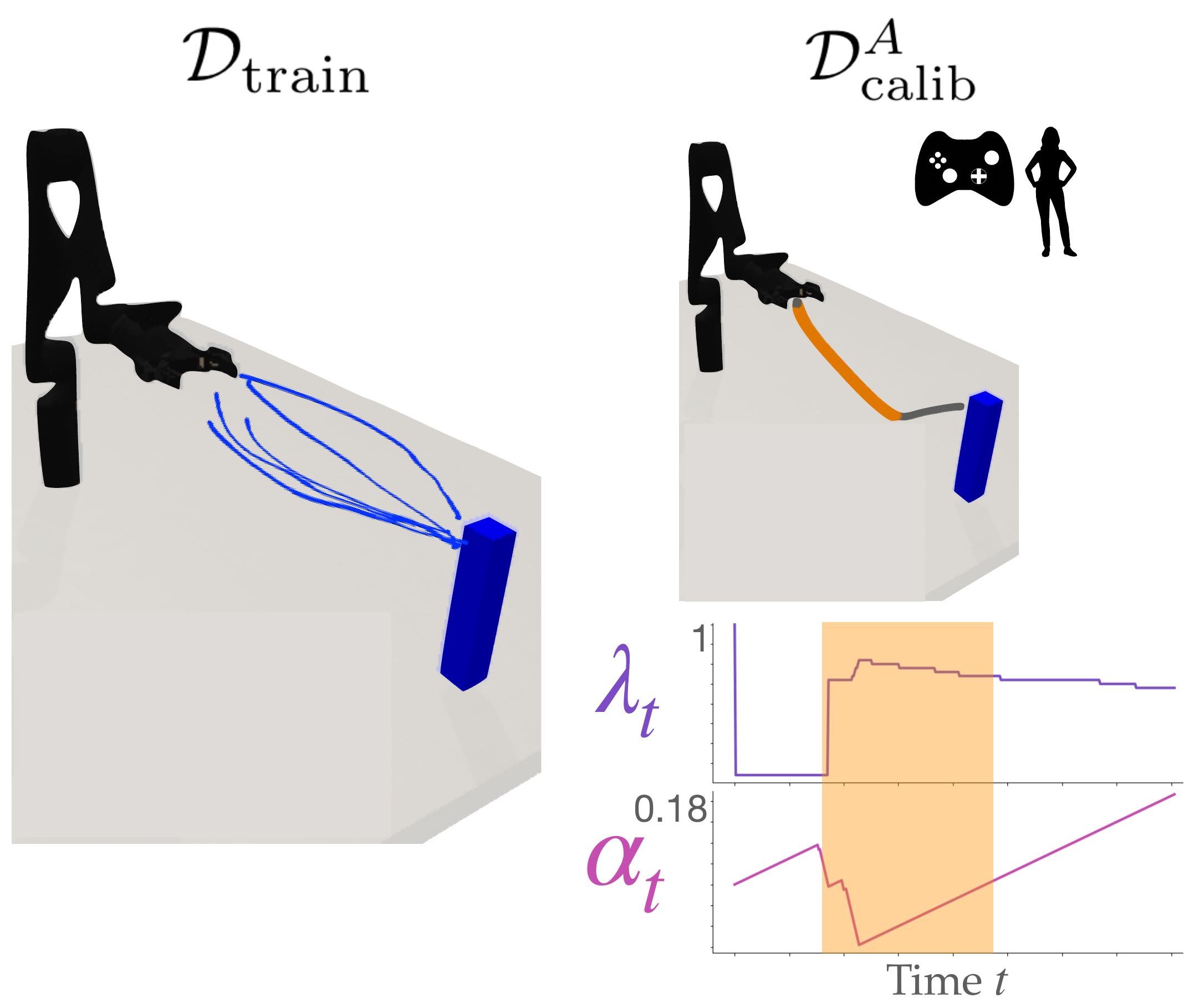}
    \caption{A single demonstrator provides demonstrations to a goal (left). ACQR flags high uncertainty throughout the trajectory's movement in free space, but has the lowest uncertainty near the object (right), where high precision is exhibited in the training data.}
    \label{goal_train_on_one}
\end{figure}

The training dataset $\dtrain$ contains 60 kinesthetically demonstrated trajectories, by a single user Alice, where all demonstrations move the end effector to the blue goal. The trajectories are very precise near the object, but less precise in the path towards the object where there's more free space. Alice employs a heuristic strategy to deterministically generate $\lowH^t$ for consecutive state pairs, where $\lowH^t$ is the change in $x$-direction and $y$-direction of the end effector from $s_t$ to $s_{t+1}$. We calibrate on the same target user, Alice, $\dcalib^A$, who demonstrates 6 unseen test trajectories of reaching the blue goal. ACQR flags high uncertainty ($\beta=0.05$) throughout the trajectory, but has the lowest uncertainty at the end near the object  (Figure \ref{goal_train_on_one}).

\para{Aside} It's important to recognize that $\alpha_t$ does not necessarily increase monotonically as a measure of uncertainty. $\alpha_t$ is the empirical quantile of the set of expansion factors needed to reach approximately $1-\alpha_t$ coverage. If the model mispredicts repeatedly, the quantiles need to be expanded by increasingly large factors, $\lambda_t$, in order to cover the true value. As $\lambda_t$, continue to be appended to set $S_{t-1}$, the set $S_t$ increasingly contains large expansion factors. As a result, the $\alpha_{t}$-adjusted quantiles do not miscover due to the present of large $\lambda_t$ in $S_t$. Thus, $\alpha_{t+1}$ may decrease while the size of the intervals (uncertainty) stays high.

\subsection{\edit{Further Results Supporting} Takeaway 4: Our proposed detection mechanism based on the uncertainty interval size is an effective way to separate high and low uncertainty states and inputs.}

\begin{table*}[t!]
\centering
{
\caption{Prediction Error in Contexts with Uncertainty Induced by Latent Preferences}
\label{tab:prediction_error_context1}
\renewcommand{\arraystretch}{1.5}
\begin{tabular}{c  c  c | c  c  c} 
 \toprule 
 \textbf{Environment} & \textbf{Target User} & \begin{tabular}{@{}c@{}} $\beta$ \end{tabular} & \begin{tabular}{@{}c@{}} $\aHhat$ \textbf{Error at Uncertain Inputs}: \edit{mean, (std)} \end{tabular} & \begin{tabular}{@{}c@{}} $\aHhat$ \textbf{Error at Certain Inputs} \end{tabular} & \textbf{T-test} \\ 
 [0.5ex] 
 \hline
 GridWorld & $\DD^A_{calib}$ & 1.0 &  0.945, ($<0.001$) & 0.0725, (0.1927) & \begin{tabular}{@{}c@{}}$t(754)=19.18$, $p<0.001$  \end{tabular} \\
 GridWorld & $\DD^B_{calib}$ & 1.0 & 1.055, ($<0.001$) & 0.0914, (0.27353) & \begin{tabular}{@{}c@{}}$t(754)=14.93$, $p<0.001$  \end{tabular}  \\ \hline 
 7DOF Goal & $\DD^A_{calib}$ & 0.05 & 0.007, (0.002) & 0.006, (0.002) & \begin{tabular}{@{}c@{}}$t(5627)=26.55$, $p<0.001$  \end{tabular} \\
 7DOF Goal & $\DD^B_{calib}$ & 0.05 & 0.010, (0.006) & 0.007, (0.003) & \begin{tabular}{@{}c@{}}$t(9655)=39.99$, $p<0.001$  \end{tabular}  \\ \hline 
 7DOF Cup & $\DD^A_{calib}$ & 0.15 & 0.014, (0.009) &  0.009, ( 0.007) & \begin{tabular}{@{}c@{}}$t(4497)=20.10$, $p<0.001$  \end{tabular}  \\
 7DOF Cup & $\DD^B_{calib}$ & 0.15 &  0.029, (0.007) & 0.009, (0.008) & \begin{tabular}{@{}c@{}}$t(2041)=61.21$, $p<0.001$  \end{tabular} \\
 \bottomrule
\end{tabular}
}

\end{table*}

In the \textbf{latent preferences} context, we also find that our mechanism separates low and high uncertainty inputs in a statistically significant manner ($p < 0.05$), and the mean prediction error is higher in uncertain states than certain states (Table \ref{tab:prediction_error_context1}). 

\begin{table*}[htbp!]
\centering
{
\caption{Latent Preferences: Coverage and Interval Size for $\dcalib^A$}
\label{tab:baselines_context_1_A}
\renewcommand{\arraystretch}{1.5}
\begin{tabular}{c  c  c  c  c  c  c} 
\toprule 
 \multicolumn{1}{c}{\textbf{Algorithm}} &  \multicolumn{3}{c}{\textbf{Coverage}} & \multicolumn{3}{c}{\textbf{Interval Length}}\\
\hline 
{} &  Grid & Goal & Cup &  Grid & Goal & Cup \\
{} &  $\dcalib^A$ & $\dcalib^A$ & $\dcalib^A$ & $\dcalib^A$ & $\dcalib^A$ & $\dcalib^A$ \\
 [0.5ex] 
 \hline
 \acqr  &   0.929 &   0.919  & 0.926 & 0.226 & 0.018 &  0.037  \\
\ensemble & 0.571 & 1.0  & 1.0  & 0.692 & 1.675  & 0.614 \\
\qr   &   0.833  &  0.821  & 0.527  & 0.214 & 0.016 & 0.016 \\
 \bottomrule
\end{tabular}
}

\end{table*}

\begin{table*}[htbp!]
\centering
{
\caption{Latent Preferences: Coverage and Interval Size for $\dcalib^B$}
\label{tab:baselines_context_1_B}
\renewcommand{\arraystretch}{1.5}
\begin{tabular}{c  c  c  c  c  c  c} 
\toprule 
 \multicolumn{1}{c}{\textbf{Algorithm}} &  \multicolumn{3}{c}{\textbf{Coverage}} & \multicolumn{3}{c}{\textbf{Interval Length}}\\
\hline 
{} &  Grid & Goal & Cup &  Grid & Goal & Cup \\
{} &  $\dcalib^B$ & $\dcalib^B$ & $\dcalib^B$ & $\dcalib^B$ & $\dcalib^B$ & $\dcalib^B$ \\
 [0.5ex] 
 \hline
\acqr  &  0.929 &  0.902  & 0.883  & 0.199  & 0.028 & 0.055  \\
\ensemble   & 0.619 & 1.0 & 1.0 & 0.697 & 1.672 & 0.653 \\
\qr   &  0.833  &   0.688 & 0.404 & 0.189 &  0.017 & 0.019\\
 \bottomrule
\end{tabular}
}

\end{table*}

\subsection{Coverage and Interval Size}

In the \textbf{latent preferences} context, \acqr achieves approximately the desired coverage on $\dcalib^A$ (Table \ref{tab:baselines_context_1_A}) and $\dcalib^B$ (Table \ref{tab:baselines_context_1_B}). On $\dcalib^A$, \acqr achieves coverage close to \ensemble with a mean interval length less than or approximate to the length of \ensemble. The $\dcalib^B$ user in \textbf{7DOF Goal-Reaching} are deliberately noisy, and the $\dcalib^B$ user in \textbf{7DOF Cup-Grasping} chooses only to grasp by the lip, which requires greater reconfiguration and thus more noisy than the handle grasp. As a result, the mean interval lengths needed by \acqr are larger than \ensemble for these two environments with $\dcalib^B$ (Table \ref{tab:baselines_context_1_B}). \qr does not achieve sufficient coverage, indicating the need to ensure calibration is performed to the end user.

In the \textbf{low-control precision} context, \acqr achieves approximately the desired coverage on $\dcalib^A$ (Table \ref{tab:baselines_context_2_A}). On $\dcalib^A$, \acqr achieves coverage close to \ensemble with a mean interval length less than that of \ensemble. \qr does not achieve sufficient coverage.

\subsection{\edit{Further Results Supporting} Takeaway 5: With human annotators providing varied, OOD low-dim input labels, \acqr ensures coverage over \qr, and our uncertainty monitoring mechanism identifies higher uncertainty on users providing out-of-distribution inputs.}
\label{subsec:red_goal_low_dim_variance}

\begin{figure}[tb]
    \centering
    \includegraphics[width = 0.49\textwidth]{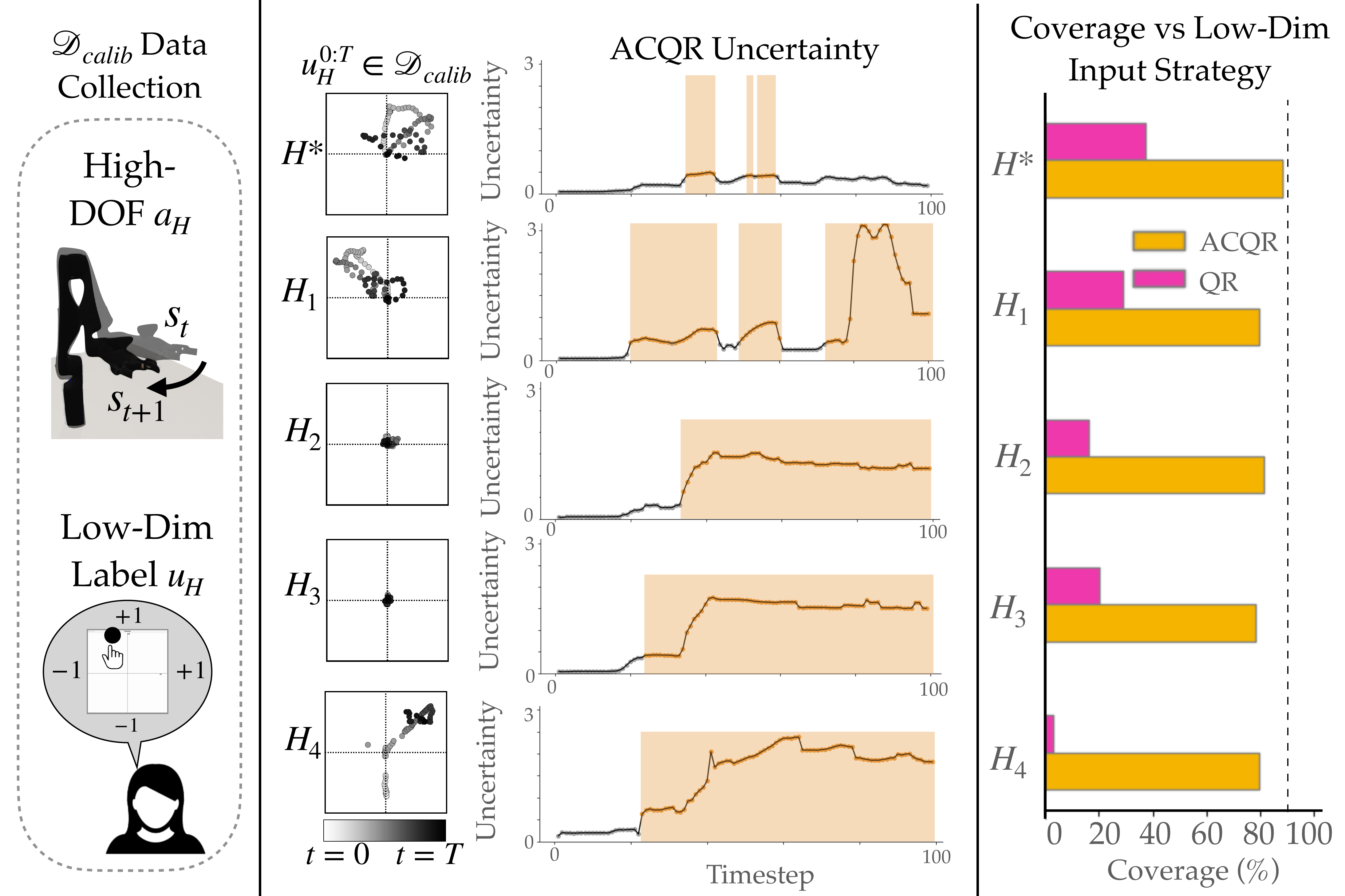}
    \caption{\edit{(Left) \textbf{7DOF goal-reaching (red) with out-of-distribution low-dimensional input schemes.} The base model without calibration, \qr, achieves less than 40\% percent coverage on both in-distribution and out-of-distribution input schemes ($H_0$-$H_4$). Calibration using \acqr increases coverage on the calibration data of all users, increasing coverage towards, but not yet at, the desired level of 90\%.} \edit{(Right) Our uncertainty monitoring mechanism, at the threshold $\beta=0.4$, flags high uncertainty for these out-of-distribution input schemes. The uncertainty tracking is able to self-assess its inability to correctly map the intended high-DOF actions when users give inputs in a way the base teleoperation controller has not been trained on.}}
    \label{fig:red_coverage}
\end{figure}

\edit{Our third uncertainty context examines variance as a result of different \textit{low-dimensional input schemes} employed by target users. We highlight this takeaway in the setting of \textbf{7DOF Goal-reaching}. We fix the base model and calibration trajectory. For the base model, we use the model trained from the setting of \textbf{7DOF Goal-reaching} with diverse \textbf{latent preferences}. Training demonstrators do not manually provide low-dimensional inputs, but instead employ a single heuristic strategy to deterministically generate $\lowH$ for consecutive state pairs, using the change in $x$-direction and $y$-direction of the end effector. 

In the calibration phase, target users provide low-dimensional input labels for two calibration trajectories: one to the blue goal, one to the red goal. We will focus this section on results for the calibration trajectory to the red goal. User $H_0$ gives input $\lowH$ equal to the change in $x$-direction and $y$-direction of the end effector, representing an in-distribution low-dimensional input scheme. Next, user $H_1$ gives input $\lowH$ equal to the change in $z$-direction and $y$-direction of the end effector, representing a slightly out-of-distribution low-dimensional input scheme. Users $H_2, H_3, H_4$ represent the three human target users (see their inputs in Figure \ref{fig:red_coverage}).

In Figure \ref{fig:red_coverage} (left), we see that the base model without calibration, \qr, achieves less than 40\% percent coverage on both the in-distribution user $H_0$ and out-of-distribution input schemes ($H_1$-$H_4$). Calibration using \acqr increases coverage on the calibration data from all in- and out-of-distribution users, increasing coverage towards the desired level of 90\%. 

In Figure \ref{fig:red_coverage} (right), our uncertainty monitoring mechanism, at the threshold $\beta=0.4$, which we increased from 0.05 for noisy human inputs, flags high uncertainty for these out-of-distribution input schemes. The uncertainty tracking flags high uncertainty frequently for users who provide low-dimensional control inputs in out-of-distribution ways.
}

\subsection{Human Labeling Cost of Low-Dimensional Input}
In analyzing uncertainty in the \textbf{low-dimensional input schemes} context, we tested realistic variance in the notions of suitable low-dimensional across different users. We collected data from three human operators, providing labels on pairs of robot states throughout two calibration trajectories. The calibration trajectory to the blue goal contained 35 timesteps, and the trajectory to the red goal contained 100 timesteps. Human operators labeled the robot action taken at each timestep. For each query, operators were able to replay the robot action as many times as needed to determine a low-dimensional input label. Operators provided the label through a graphical interface, where they clicked on a screen to provide what would represent their precise joystick input (see Figure \ref{fig:blue_coverage} (left)). Each annotation session was timed, and the annotation time was averaged over each timestep across both trajectories to determine an average labeling cost in seconds per query (Table \ref{tab:labeling_cost}). As a result, these labeling costs per query include the loading time of the interface. We present these labeling costs as merely an observation of the labeling cost associated with our interface. We recognize that labeling costs may differ based on the input interface, presentation of robot action, and amount of operator experience in controlling robot systems. While the cost is not too high on a per-state basis, relaxing the need for per-timestep labels for calibration is an area for future work.

\begin{table}[htbp!]
\centering
{
\caption{Labeling Cost}
\label{tab:labeling_cost}
\begin{tabular}{c c } 
\hline 
 \textbf{Human Operator} &  \textbf{Avg. Query Respond Time (s)}\\
\hline 
$H_2$ &  10  \\
$H_3$ &  8  \\
$H_4$ &  7  \\
 \hline
\end{tabular}
}
\end{table}

\end{document}